\documentclass{article}

\PassOptionsToPackage{numbers, compress}{natbib}

\usepackage{natbib}
 \usepackage[preprint]{neurips_2026}

\usepackage[utf8]{inputenc} 
\usepackage[T1]{fontenc}    
\usepackage{hyperref}       
\usepackage{url}            
\usepackage{booktabs}       
\usepackage{amsfonts}       
\usepackage{nicefrac}       
\usepackage{microtype}      
\usepackage[table]{xcolor}  

\usepackage{algorithmic}
\usepackage{wrapfig}
\usepackage[most]{tcolorbox}
\newcommand{\placeholdergraphic}[2][0.95\linewidth]{%
  \fbox{\begin{minipage}[c][0.18\textheight][c]{#1}\centering\footnotesize #2\end{minipage}}%
}
\usepackage{capt-of}    

\usepackage{amsmath}
\usepackage{amssymb}
\usepackage{amsthm}
\usepackage{enumitem}
\usepackage{multicol}
\usepackage{multirow}
\usepackage{graphicx}   
\usepackage{needspace}  
\usepackage{tabularx}
\usepackage{pifont}

\usepackage{tikz}
\usetikzlibrary{positioning,arrows.meta,calc,fit,backgrounds}

\definecolor{cGray}{HTML}{F1EFE8}
\definecolor{cGrayB}{HTML}{5F5E5A}
\definecolor{cGrayT}{HTML}{2C2C2A}

\definecolor{cPurple}{HTML}{EEEDFE}
\definecolor{cPurpleB}{HTML}{534AB7}
\definecolor{cPurpleT}{HTML}{26215C}

\definecolor{cTeal}{HTML}{E1F5EE}
\definecolor{cTealB}{HTML}{0F6E56}
\definecolor{cTealT}{HTML}{04342C}

\definecolor{cCoral}{HTML}{FAECE7}
\definecolor{cCoralB}{HTML}{993C1D}
\definecolor{cCoralT}{HTML}{4A1B0C}

\newcommand{\Dcal}{\mathcal{D}}

\newcommand{\Scal}{\mathcal{S}}

\newcommand{\Score}{\mathrm{Score}}
\newcommand{\err}{\mathrm{err}}

\title{Agentic Search Spaces for Tabular Machine Learning}

\author{%
  Renat Sergazinov$^{\dagger}$\thanks{Corresponding author, email at <mc.sergazinov@gmail.com> $^\dagger$ Yandex $^\ddagger$ HSE University} \\
  \And
  Artem Chistyakov$^{\dagger \ddagger}$ \\
  \And
  Sergey Pankevich$^{\dagger \ddagger}$ \\
  \And
  Artem Babenko$^{\dagger\ddagger}$ \\
}

\begin{document}

\maketitle

\begin{abstract}

Despite the rapid progress of LLM-based agents for planning, code generation, and debugging, their practical value for tabular machine learning remains underexplored. In this paper, we investigate a concrete use case: whether state-of-the-art agentic AI systems can design extended HPO search spaces for established tabular models that outperform the standard search spaces provided by the model authors.

Specifically, we represent each tabular model as a modular pipeline covering preprocessing, embeddings, architecture, training, and inference. We then task the agent to propose candidate code implementations for each module and use a classical HPO algorithm to jointly optimize over these candidates and the model's default hyperparameters. Compared with the base HPO spaces, the expanded search spaces improve the performance of nearly every model family across a suite of 45 datasets, with average relative gains of $0.6\%$, rising to $2.0\%$ on small-to-medium regression datasets. Notably, these gains come at no extra tuning cost: the enlarged spaces outperform the base under the same tuning and ensembling budgets. The gains transfer to the recent TabArena benchmark, where the agentic spaces improve the official Elo scores of four of the five model families and the two strongest agentic ensembles surpass the best AutoGluon ensemble of conventional models. Overall, our study suggests that LLM agents can provide practical value for tabular ML by expanding the design space. The code is available at: \url{https://github.com/yandex-research/agentic-hpset}.

\end{abstract}

\section{Introduction}
\label{sec:introduction}

Tabular data is one of the most abundant modalities in machine learning, with applications across finance, healthcare, recommender systems, and scientific data analysis. This practical importance has made tabular ML an active research area, with recent progress in specialized DNN architectures \citep{holzmueller2024realmlp, gorishniy2025tabm, ye2025modernnca}, foundational models \citep{hollmann2023tabpfn,qu2025tabicl,zhang2025limix}, and more rigorous benchmarks and evaluation protocols \citep{rubachev2024tabred,erickson2025tabarena}.

At the same time, the recent rise of large language models (LLMs) has so far had only limited impact on state-of-the-art tabular ML. While prior work has investigated LLMs for direct
prediction over serialized tables~\citep{hegselmann2023tabllm,gardner2024tabula},
semantic feature engineering~\citep{hollmann2023caafe,han2024featllm}, and
prediction with textual metadata~\citep{grinsztajn2023modeling, yakushev2025talking}, these approaches appear most useful in specific niches, e.g.,\ when datasets are small or feature names are
semantically informative. For more general cases, however, LLMs have not substantially affected standard practitioner workflows in tabular ML.

\begin{figure}[t]
\centering
\resizebox{\linewidth}{!}{%
\begin{tikzpicture}[
    font=\sffamily\small,
    >={Stealth[length=2mm,width=1.6mm]},
    every box/.style={
        rounded corners=2pt,
        draw, line width=0.4pt,
        align=center, inner sep=4pt,
        text width=24mm, minimum height=11mm,
    },
    gray box/.style={every box, fill=cGray, draw=cGrayB, text=cGrayT},
    purple box/.style={every box, fill=cPurple, draw=cPurpleB, text=cPurpleT},
    teal box/.style={every box, fill=cTeal, draw=cTealB, text=cTealT},
    coral box/.style={every box, fill=cCoral, draw=cCoralB, text=cCoralT},
    arrow/.style={->, line width=0.5pt, draw=cGrayB},
    stage label/.style={font=\sffamily\small\bfseries, text=cGrayT},
]
 
\node[gray box] (model) at (0, 0) {Base model \\ \footnotesize\itshape MLP, TabM, GBDT};

\node[purple box, right=10mm of model.east,
      text width=22mm, minimum height=18mm] (agent)
      {LLM agent \\ \footnotesize\itshape Generates per module};

\draw[arrow] (model.east) -- (agent.west);
 
\node[teal box, right=10mm of agent.east,
      text width=30mm, minimum height=32mm,
      inner sep=3pt] (hbox) {%
    \footnotesize\itshape Candidate sets \\[1pt]
    \footnotesize
    $\mathcal{H}_1$ \, preprocessing \\
    $\mathcal{H}_2$ \, embedding \\
    $\mathcal{H}_3$ \, architecture \\
    $\mathcal{H}_4$ \, training \\
    $\mathcal{H}_5$ \, inference%
};
 
\draw[arrow] (agent.east) -- (hbox.west);
 
\node[teal box, right=12mm of hbox.east,
      text width=28mm, minimum height=14mm] (space)
      {$\mathcal{H}_1 \!\times\! \cdots \!\times\! \mathcal{H}_5 \!\times\! \mathcal{S}_{\mathrm{HPO}}$
       \\ \footnotesize\itshape Joint search space};
 
\node[purple box, right=8mm of space.east,
      text width=22mm, minimum height=14mm] (hpo)
      {Bayesian HPO \\ \footnotesize\itshape Selects joint config};
 
\node[gray box, below=8mm of space.south,
      text width=24mm, minimum height=11mm] (data)
      {New dataset \\ \footnotesize\itshape Train/val splits};
 
\node[coral box, right=8mm of hpo.east,
      text width=22mm, minimum height=14mm] (final)
      {Fitted pipeline \\ \footnotesize\itshape Best joint config};
 
\draw[arrow] (hbox.east)  -- (space.west);
\draw[arrow] (space.east) -- (hpo.west);
\draw[arrow] (data.north) -- (space.south);
\draw[arrow] (hpo.east)   -- (final.west);
 
\node[stage label] at ($(agent.south)+(0,-3.0)$)
    {Stage 1 --- per model (run once)};
\node[stage label] at ($(hpo.south)+(0,-3.0)$)
    {Stage 2 --- per dataset (run each time)};
 
\draw[dashed, line width=0.3pt, draw=cGrayB]
    ($(hbox.north east)+(6mm,4mm)$) -- ($(hbox.south east)+(6mm,-25mm)$);
\end{tikzpicture}
}
\label{fig:main-diag}
\caption{Method overview. \emph{Stage 1} runs once per base model: an LLM agent generates candidate sets $\mathcal{H}_1, \dots, \mathcal{H}_5$ for the five pipeline modules. \emph{Stage 2} runs on each new dataset: a classical HPO searches the joint space and returns a fitted pipeline.}
\vspace{-1em}
\end{figure}
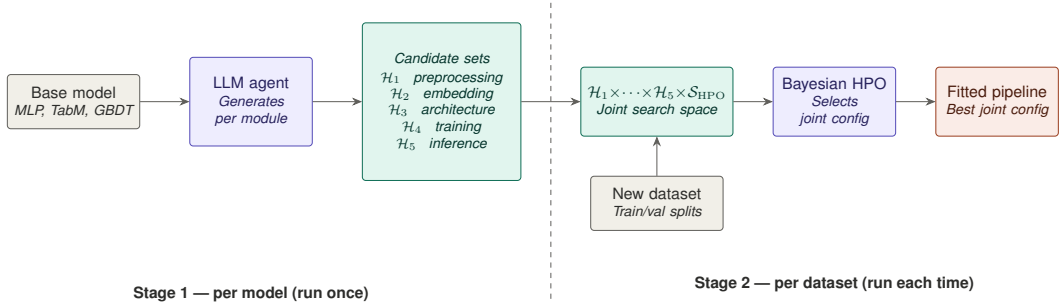

Recent developments in agentic AI systems suggest new possible roles for LLMs in tabular problems.
In particular, the ability of advanced AI agents to analyse and improve code is well suited for an under-automated part of tabular ML practice: \emph{identifying which recent techniques or combinations of them can improve a particular model}. Most tabular models are released as a frozen recipe with a small hyperparameter (HPO) set. In practice, however, users frequently hand-tune models based on domain knowledge and current literature trends (e.g. new activation functions or an optimizer) with no systematic way to explore different options or their interactions. In this paper, we investigate whether state-of-the-art agents can close this gap automatically. Given a model, we split it into a sequence of code modules (e.g. preprocessing, embedding, architecture) and task the agent to propose and implement in code the alternatives for each of them. These code modules are then added as categorical variables to the HPO space. We visualize our proposed method in Figure~\ref{fig:main-diag}. The model is then tuned over this extended search space using a standard hyperparameter optimization (HPO). This use of LLMs is attractive for several reasons.  First, it improves performance: we find the enlarged search space lets HPO find model variants that outperform the model tuned over the default author-provided HPO space. Second, it simplifies model upkeep: practitioners no longer need to find and test each new idea manually. Third, the cost of the LLM can be amortized across the datasets: the search space can be generated once per-model without access to the target data.

We evaluate agent-augmented HPO search spaces across representative tabular model families, including DNNs, GBDTs, and tabular foundation models, on a curated set of 45 datasets. Compared to the default HPO spaces, the agent-augmented spaces improve the aggregate rank of nearly every family. Averaged across all datasets, tuned single models gain $0.5\%$ for DNNs (MLP$^\dagger$, TabM$^\dagger$, RealMLP), $0.8\%$ for GBDTs (LightGBM), and $0.5\%$ for the TabICLv2 foundation model, while ensembles gain $0.9\%$ for DNNs and $0.2\%$ for GBDTs. The gains concentrate on small-to-medium regression datasets, where the average single-model gains rise to $2.0\%$ for DNN and GBDT, while TabICLv2 improves on large regression datasets by $2.1\%$. The richer spaces also improve hyperparameter ensembling, reducing prediction correlation and yielding stronger ensembles across the board. Finally, the conclusions transfer to the recent TabArena benchmark, where the agentic spaces improve the Elo scores of four of the five model families, by up to $125$ points.

Overall, our results suggest a practical and complementary role for LLM agents in tabular ML. Rather than replacing specialized tabular models or performing expensive dataset-specific experimentation, agents can act as automated search-space designers that transfer broad ML implementation knowledge into reusable, model-specific HPO spaces.

The contributions of our paper are the following:

\begin{enumerate}[leftmargin=2em]

\item We propose a new method for improving tabular models with LLM agents: the researcher splits the model into a sequence of modules, the agent generates candidate module implementations, and a traditional HPO algorithm explores the resulting joint search space on each dataset. We make our prompts and harnesses publicly available.
\item We test the method by generating extended HPO search spaces, which we also release, for five model families: MLP$^\dagger$, TabM$^\dagger$, LightGBM, RealMLP, and TabICLv2.
\item On these models, we show consistent average gains across the 45 datasets: $0.5\%$ for DNNs, $0.8\%$ for GBDTs, and $0.5\%$ for TabICLv2 in the single-model regime, and $0.9\%$ for DNNs in the ensemble regime with up to $2.0\%$ on small-to-medium regression tasks. The ensemble over the agentic RealMLP space is the strongest of all model families in our study, outperformed only by the multi-model AutoGluon system. We also evaluate on the TabArena benchmark and observe consistent gains there as well.
\end{enumerate}

\section{Related Works}

\textbf{Tabular deep learning and pipeline design.}
Recent tabular work has advanced both architectures (TabM~\citep{gorishniy2025tabm}, TabR~\citep{gorishniy2024tabr}, ModernNCA~\citep{ye2025modernnca}) and foundation models (TabPFN~\citep{hollmann2023tabpfn}, TabICLv2~\citep{qu2026tabiclv2}, LimiX~\citep{zhang2025limix}), while benchmarks like TabReD~\citep{rubachev2024tabred} and TabArena~\citep{erickson2025tabarena} show that protocol choices substantially affect conclusions.

\textbf{AutoML and architecture search.} Auto-sklearn~\citep{feurer2015autosklearn} builds a greedy ensemble~\citep{caruana2004ensemble} over combinations of preprocessing methods and shallow models. AutoGluon~\citep{erickson2020autogluon} follows the same ensemble-first recipe, but does not optimize over preprocessing and broadens the model library from shallow learners to deep architectures (e.g.\ TabM \citep{gorishniy2025tabm}, RealMLP \citep{holzmueller2024realmlp}) and foundation models (e.g.\ TabICLv2 \citep{qu2026tabiclv2}, LimiX \citep{zhang2025limix}). AutoPyTorch~\citep{zimmer2021autopytorch} instead ensembles parametrizations of a single model family, varying regularizers, activations, and MLP layer shapes. AutoKeras~\citep{jin2019autokeras} similarly searches over neural architectures assembled from its own library of implemented components~\citep{zoph2017nas,liu2019darts}. Our setting is closest to AutoPyTorch and AutoKeras: we start from a single model and search over its modifications. We differ in two key ways. First, the candidate generation is conditioned on the input model: the agent proposes modules specifically for the given model. Second, the candidates are written by the agent rather than hand-implemented. The resulting models with extended HPO spaces can then be wrapped and used with Auto-sklearn and AutoGluon.

\textbf{LLMs for tabular prediction.} Prior LLM-based approaches to tabular prediction fall into two families. The first treats the LLM as a direct predictor over serialized rows, via prompting (TabLLM~\citep{hegselmann2023tabllm}) or fine-tuning (Tabula-8B~\citep{gardner2024tabula}); the second uses the LLM for semantic feature engineering, generating new columns from dataset descriptions (CAAFE~\citep{hollmann2023caafe}, FeatLLM~\citep{han2024featllm}). Both have shown gains in regimes such as small datasets or rich textual metadata.

\textbf{LLM agents for optimization and research automation.} Prior work uses LLMs either as optimizers~\citep{liu2024llambo,chen2022optformer,meindl2025gptopt,menet2025tosfit,tan2025offlinebbo} or as agents that automate parts of the research loop~\citep{lu2024aiscientist,karpathy2026autoresearch,ferreira2026can}. The former is orthogonal to our work: these methods change how a fixed search space is explored, whereas we use a standard sampler and change the space itself. Our work is closer to the latter, but restricts the agent to generating modules along predefined pipeline axes, followed by controlled HPO. This preserves dataset privacy, amortizes agent cost, and makes the resulting search spaces reproducible and amenable to ablation.

\section{Methodology}
\label{sec:methodology}

\subsection{Setup and notation}
\label{sec:setup}

\textbf{Datasets.} We consider a tabular task with a fixed split $\Dcal=(\Dcal_{\mathrm{train}},\Dcal_{\mathrm{val}},\Dcal_{\mathrm{test}})$. We use datasets derived from TabM \citep{gorishniy2025tabm}, TabArena \citep{erickson2025tabarena}, and TabReD \citep{rubachev2024tabred} benchmarks, which span both regression and classification datasets ranging from
768 to 1M+ objects and from 5 to 1500+ features. We report their summary statistics in Table~\ref{tab:dataset_statistics} and list their full properties in Appendix~\ref{app:setup-details-data}. Our design choices for the benchmark were to: (1) include both i.i.d.\ and non-i.i.d.\ datasets; (2) keep a roughly balanced split between regression and classification; (3) include datasets with a diverse range of sample-to-feature ratios. In addition, we separately report our model performance on the TabArena benchmark in Appendix~\ref{app:tabarena}.

\begin{table}[h]
\caption{Overview of the datasets included in the main evaluation; sizes are total row counts. Full per-dataset statistics are in Appendix~\ref{app:setup-details}.}
\label{tab:dataset_statistics}
\scriptsize
\setlength{\tabcolsep}{3.0pt}
\begin{center}
\begin{tabular}{cccccccccccc}
\toprule
\multicolumn{2}{c}{Source}
& \multicolumn{4}{c}{\#Rows}
& \multicolumn{4}{c}{\#Features}
& \multicolumn{2}{c}{Task type} \\
\cmidrule(lr){1-2}
\cmidrule(lr){3-6}
\cmidrule(lr){7-10}
\cmidrule(lr){11-12}
OpenML & TabReD
& Min. & Q50 & Mean & Max.
& Min. & Q50 & Mean & Max.
& \#Regr. & \#Classif. \\
\midrule
37 & 8
& 768 & 21K & 103K & 1.2M
& 5 & 21 & 156 & 1776
& 20 & 25 \\
\bottomrule
\end{tabular}
\end{center}
\end{table}

\textbf{Metrics.} We evaluate every dataset with its native metric adopted from the source benchmark (AUROC, accuracy, or log-loss for classification and RMSE for regression); the per-dataset metrics are listed in Appendix~\ref{app:setup-details-data}. We report four aggregate metrics, all at a fixed tuning budget $B$: mean rank, relative improvement over the tuned base-space MLP$^\dagger$, normalized score, and Elo score. We define these metrics in Appendix~\ref{app:setup-details-metrics}.

\textbf{Models and agents.} We select strong within-family implementations. To represent DNNs, we take, in order of increasing complexity: (i) MLP$^\dagger$ -- an MLP with piecewise-linear embeddings of \citet{gorishniy2022embeddings}; (ii) TabM$^\dagger$ -- the mini variant of TabM with piecewise-linear embeddings and $k=32$ members from \citet{gorishniy2025tabm}; (iii) RealMLP -- the MLP model from PyTabKit with the latest TabArena hyperparameters \citep{holzmueller2024realmlp, erickson2025tabarena}.  To represent GBDT, we take the LightGBM model \citep{ke2017lightgbm}. Among foundational models, we take the recently proposed TabICLv2 \citep{qu2026tabiclv2}; since it performs no gradient-based training, we refit it on the concatenated train and validation splits to obtain test predictions (Appendix~\ref{app:setup-details-hps}).  As an agent, we use Claude Code running Opus 4.8 on \texttt{max} effort settings \citep{anthropic2026claudecode, anthropic2026opus48}. We additionally run speed and performance ablations with Codex running GPT 5.5 with \texttt{extra-high} settings \citep{openai2025codex, openai2026gpt55}.

\textbf{Evaluation setup.} We tune each model over its HPO space with Optuna's univariate Tree-structured Parzen Estimator (TPE) sampler~\citep{bergstra2011algorithms,akiba2019optuna}. The first $20$ trials are sampled at random to seed the densities. Each trial trains one configuration on the training subset and scores it on the validation subset; the budget is $200$ trials on small and medium datasets and $100$ on the large ones (Microsoft and TabReD), after which the configuration with the best validation score is evaluated on the held-out test subset with 15 random seeds. TabICLv2 is the exception: the default model is used as published, without tuning, while its agentic variant is tuned with a fixed budget of $100$ trials on every dataset. For the ensembling study, we build greedy ensembles following \citet{caruana2004ensemble}: we sample $100$ random configurations per search space ($20$ on large datasets), train them, and iteratively add, with replacement, the model whose inclusion in the prediction average most reduces the validation error. The selected ensemble is then evaluated on the test subset; further details on the greedy ensemble algorithm are given in Appendix~\ref{app:setup-details-hpo}.

\subsection{From end-to-end agents to agent-generated search spaces}
\label{sec:proposed}

\begin{table*}[t]
\caption{Selected agent-proposed candidate implementations per module. Tags \texttt{(reg.)}, \texttt{(clf.)}, \texttt{(cat.)}, and \texttt{(num.)} mark candidates applicable only to regression, classification, categorical features, or numerical features, respectively. The trailing number in parentheses is the count of agent-generated candidates for that module; full per-family lists are in Appendix~\ref{app:llm-skills-and-guardrails}.}
\label{tab:agent_modules_select}
\scriptsize
\setlength{\tabcolsep}{2.2pt}
\renewcommand{\arraystretch}{1.05}
\begin{center}
\begin{tabularx}{\textwidth}{@{}l*{5}{>{\raggedright\arraybackslash}X}@{}}
\toprule
Model & Preprocess / data aug. & Embedding (cat.+num.) & Train & Inference / eval. & Architecture \\
\midrule
MLP$^\dagger$ &
Yeo--Johnson, RankGauss, winsorized, row-stats (num.); rare-bucket / target-order (cat.); symlog / quantile target (reg.) (14) &
PLR/PLE, B-spline, periodic, RFF, bilinear (num.); entity / hashed / target-mean (cat.)\ \citep{gorishniy2022embeddings} (21) &
AdamW (cosine / warm restarts), EMA~\citep{polyak1992acceleration}, SWA, model soup, MixUp, Lookahead; Huber / Gaussian-NLL (reg.); label smoothing (clf.) (15) &
Train-range clipping (reg.); MC-dropout, Gaussian-noise TTA, temperature scaling (clf.) (5) &
Bottleneck, feature-dropout, GELU/GLU, residual, pre-norm, SE, wide--narrow (12) \\
\bottomrule
\end{tabularx}
\end{center}
\end{table*}

The most direct way to hand a tabular model to an agent is to let it edit the model end to end: starting from the reference implementation, the agent changes a hyperparameter or a piece of code, runs a validation fit, and keeps the change if the score improves. This is known as the autoresearch loop described in \citet{karpathy2026autoresearch}, which we adapt to the tabular setting and run for $50$ iterations on four datasets, starting from the $\mathrm{MLP}^\dagger$ code (Appendix~\ref{app:autore-prog}). Table~\ref{tab:autoresearch_motivation} summarizes the outcome: the loop costs \$20--50 \emph{per dataset}, requires handing the agent the data, and ends up marginally behind classical HPO over the default HPO space.

\begin{table}[h]
\caption{The end-to-end autoresearch loop~\citep{karpathy2026autoresearch} against classical HPO and our method, on the four datasets of Table~\ref{tab:autoresearch_vs_hpo} at a matched budget of $50$ trials. $\Delta$ is the mean relative improvement.}
\label{tab:autoresearch_motivation}
\scriptsize
\setlength{\tabcolsep}{6pt}
\begin{center}
\begin{tabular}{llllr}
\toprule
Method & Agent is run & Agent cost & Sees the data & $\Delta$ over classical HPO \\
\midrule
Autoresearch~\citep{karpathy2026autoresearch} & per dataset & \$20--50 per dataset & yes & $-0.1\%$ \\
Ours & per model  & \$10 per model & no & $+1.3\%$ \\
\bottomrule
\end{tabular}
\end{center}
\end{table}

\textbf{Three problems.} We attribute this negative result to three properties of the end-to-end loop.

\emph{(i) Few ideas generalize well in tabular data; diversity and number are therefore an advantage.} The winning module combinations are pairwise distinct across the 18 datasets we examine (Section~\ref{sec:module-importance}), and expanding the candidate pool monotonically grows performance (Section~\ref{sec:candidate-pool}). No single module generalizes well across the datasets, so the search should maximize the \emph{diversity} and \emph{number} of explored ideas, covering as much of the idea space as possible. Contrary to this, the end-to-end loop tends to converge on a narrow set of techniques. Indeed LLMs are known to explore poorly in-context \citep{krishnamurthy2024can}, and agents in long loops increasingly revisit earlier actions as their context fills up \citep{wang2026notime, xu2026loopguard}.

\emph{(ii) The search space is explored inefficiently.} Even for the ideas it does try, the greedy LLM-based search is a poor search strategy: it evaluates one edit at a time, accepts greedily, does not maintain a model of the response surface, and proposes candidates based purely on the LLM prior. The weakness is sharpest for the numerical hyperparameters, which hold a large share of the achievable gain (learning rate, weight decay; Figure~\ref{fig:module_lift}, blue boxes). On exactly such problems, classical black-box optimization algorithms such as TPE sampler (implemented in Optuna) outperform LLMs \citep{ferreira2026can}.

\emph{(iii) The cost does not amortize across datasets.} The loop is re-run for every new dataset, so its price scales with the benchmark. Repeating our 45-dataset study this way would cost at least \$900--\$2{,}300 in agent calls alone. The agent also needs the data to score its own edits, which invites contamination.

\textbf{Three fixes.} Our method addresses each problem in turn.

\emph{(i) Steer the agent towards coverage.} We decompose each model by hand into a diverse set of mutually non-overlapping modules supported by the literature. For the neural families (MLP$^\dagger$, TabM$^\dagger$, and RealMLP), we take the modules to be preprocessing, feature embeddings, architecture, training, and inference with the default taken from the original method, e.g., for MLP$^\dagger$ quantile-transform as numerical preprocessing, AdamW \citep{loshchilov2019adamw} as optimizer. For LightGBM and TabICLv2, whose learning algorithms are frozen, we take the modules to be preprocessing and inference. The agent proposes alternative implementations of each module as self-contained code blocks, e.g., Yeo--Johnson feature transform \citep{yeo2000power}, Lion optimizer \citep{chen2023lion}. We instruct the agent to generate at most $64$ candidates per task type (regression and classification), asking it to filter out proposals that are repetitive, do not pass the evidence bar (i.e., have been tried and failed in the literature before), or repeatedly fail the synthetic smoke tests (Appendix~\ref{app:llm-skills-and-guardrails-hypo-workflow}). Each module then enters the search space as a categorical variable with a uniform prior, e.g., $\text{train module} \sim \mathrm{Categorical}(\{\text{AdamW},\, \text{AdamW + EMA},\, \dots\})$. Our design promotes high coverage of potentially promising model configurations.

\emph{(ii) Delegate the exploration to a classical sampler.} The agent only generates module alternatives, expressed as code. The joint space of modules and continuous hyperparameters is then explored by a standard HPO sampler. Hence, the work is divided between the agent and the HPO algorithm.

\emph{(iii) Generate once per model family.} The agent generates once per model family. Concretely, it receives the model's code, the compressed title and abstract of the original paper, and tools to test candidate implementations, but no target dataset. The generation cost then amortizes over every dataset the space is later used on, and contamination or dataset leakage are excluded by construction.

Table~\ref{tab:agent_modules_select} shows a selection of the proposed modules for MLP$^\dagger$; the full sets for each model and the complete agent environment are described in Appendix~\ref{app:llm-skills-and-guardrails-hypo-workflow}.

\textbf{Alternatives.} The design above deliberately produces a single space per model, which standard HPO then adapts per dataset. One could instead generate a separate space for each dataset, or further admit conditional hyperparameters. We avoid this for the reasons in problem (iii), and because public benchmarks are mostly small to medium in size, so exponentially expanding the space with conditional hyperparameters would likely lead to overfitting. For a practitioner working on a private large application dataset, however, tailoring the space to a specific dataset may prove beneficial.

\section{Results}
\label{sec:results}

\subsection{Main performance}
\label{sec:main-performance}

\begin{figure}[t]
    \centering
    \includegraphics[width=\linewidth]{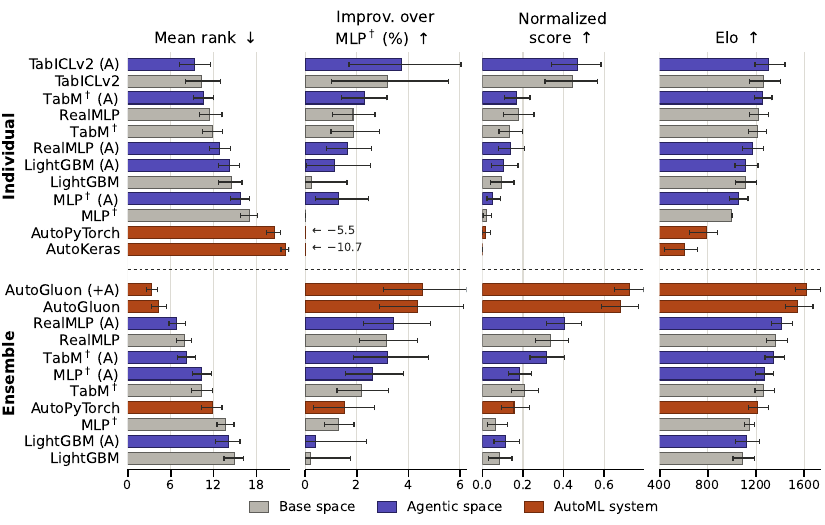}
    \caption{Aggregate metrics across 45 datasets for tuned individual models (top) and greedy ensembles (bottom). Individual models are tuned for 200 trials on small-to-medium datasets and 100 trials on large (agentic TabICLv2: 100 trials everywhere; default TabICLv2: not tuned); ensembles are built from pools of 100 (20 for large datasets) random configurations. All metrics are computed over the joint pool; improvement over MLP is the mean relative improvement on the dataset-native metrics with respect to the tuned base-space MLP$^\dagger$. Purple bars and the (A) suffix mark the agent-augmented spaces. Orange bars are AutoML systems. AutoGluon (+A) is AutoGluon with the agentic RealMLP and TabM$^\dagger$. Error bars are $95\%$ confidence intervals across datasets.}
    \label{fig:main_results}
\end{figure}

\textbf{Tuned models.} Figure~\ref{fig:main_results} (top) summarizes the tuned single-model results: in four of the five model families, the agent-augmented space improves the aggregate standing over its base space, consistently across all four metrics; the exception is RealMLP, whose base space is meta-tuned. Table~\ref{tab:grouped_improvement} breaks the effect down by task type and dataset size (see Appendix~\ref{app:setup-details-data} for the size definition). The gains are cleanest on small-to-medium regression datasets: agentic MLP$^\dagger$ improves over its base space by $+3.7\%$ on small and $+1.6\%$ on medium datasets, agentic LightGBM by $+2.1\%$ and $+1.8\%$, and agentic TabM$^\dagger$ by $+1.8\%$ and $+1.6\%$; on large regression datasets the improvements shrink to $-0.4$--$+0.1\%$. Classification is more saturated: agentic MLP$^\dagger$ still gains $+2.1\%$ on small data, while on medium datasets the DNNs and LightGBM move within $\pm0.4\%$ and on large ones the gains stay below $+0.7\%$. TabICLv2 is already the strongest single model on small datasets, leaving little room for improvement ($+0.1$--$+0.4\%$), but benefits from the agentic space exactly where it lags behind: $+2.1\%$ on large regression and $+1.1\%$ on large classification. RealMLP shows the saturation effect in its strongest form: its base space ships with meta-learned defaults, and the agentic space leaves the tuned single model essentially unchanged on regression while degrading classification by $-0.8\%$ on medium and large datasets. The main conclusion is therefore not that an agent always helps, but that agent-generated spaces help most when the base family exposes useful, composable modules and the task is not already saturated.

\begin{table}[t]
\caption{Mean relative improvement (\%) of the agent-augmented space over the corresponding base HP space, computed on the dataset-native metrics for tuned single models and greedy ensembles, grouped by task type and dataset size. Group sizes (small/medium/large): regression $8/6/6$, classification $10/12/3$ datasets. TabICLv2 is not part of the ensembling study.}
\label{tab:grouped_improvement}
\scriptsize
\setlength{\tabcolsep}{5pt}
\begin{center}
\begin{tabular}{ll rrr rrr}
\toprule
 & & \multicolumn{3}{c}{Regression} & \multicolumn{3}{c}{Classification} \\
\cmidrule(lr){3-5}\cmidrule(lr){6-8}
Method & Scope & Small & Medium & Large & Small & Medium & Large \\
\midrule
\multirow{2}{*}{MLP$^\dagger$} & Single & $+3.7$ & $+1.6$ & $+0.1$ & $+2.1$ & $-0.4$ & $+0.5$ \\
 & Ensemble & $+3.2$ & $+1.6$ & $-0.6$ & $+2.1$ & $+0.5$ & $+0.2$ \\
\midrule
\multirow{2}{*}{TabM$^\dagger$} & Single & $+1.8$ & $+1.6$ & $-0.1$ & $-0.2$ & $-0.4$ & $+0.7$ \\
 & Ensemble & $+1.5$ & $+1.5$ & $-0.4$ & $+2.9$ & $0.0$ & $+0.6$ \\
\midrule
\multirow{2}{*}{LightGBM} & Single & $+2.1$ & $+1.8$ & $-0.4$ & $+1.2$ & $-0.1$ & $+0.2$ \\
 & Ensemble & $+1.0$ & $+2.6$ & $-2.3$ & $+1.0$ & $-0.8$ & $-0.4$ \\
\midrule
\multirow{2}{*}{TabICLv2} & Single & $+0.1$ & $+0.2$ & $+2.1$ & $+0.4$ & $+0.1$ & $+1.1$ \\
 & Ensemble & -- & -- & -- & -- & -- & -- \\
\midrule
\multirow{2}{*}{RealMLP} & Single & $+0.3$ & $+0.3$ & $-0.3$ & $0.0$ & $-0.8$ & $-0.8$ \\
 & Ensemble & $+0.9$ & $-0.6$ & $+1.0$ & $+0.4$ & $+0.1$ & $0.0$ \\
\bottomrule
\end{tabular}
\end{center}
\end{table}

\textbf{Tuned and ensembled models.} In Figure~\ref{fig:main_results} (bottom), ensembling amplifies the effect: ensembles drawn from the agentic spaces occupy the top of the leaderboard, and the agentic RealMLP ensemble is the strongest of all model families, behind only the multi-model AutoGluon system (full per-dataset tables are in Appendix~\ref{app:ensemble-per-dataset}). The ensemble rows of Table~\ref{tab:grouped_improvement} follow the single-model pattern but are more uniform across task groups: agentic MLP$^\dagger$ gains $+3.2\%$ on small and $+1.6\%$ on medium regression and $+2.1\%$ on small classification, agentic TabM$^\dagger$ reaches $+2.9\%$ on small classification, and RealMLP gains $+0.9\%$ on small and $+1.0\%$ on large regression. Table~\ref{tab:ensemble_summary} explains where these gains come from. The agentic pools are only marginally stronger point-wise: the best single member of an agentic pool matches its base counterpart within $+0.4\%$ on average. Their predictions are, however, substantially less correlated. Hence, the boost from the agentic space is more pronounced for ensembles than for the single models.

\textbf{AutoML systems.} Figure~\ref{fig:main_results} also compares the search spaces against three AutoML systems run as-is on the same datasets and test splits: AutoKeras \citep{jin2019autokeras}, AutoPyTorch \citep{zimmer2021autopytorch}, and AutoGluon \citep{erickson2020autogluon}. Following \citet{gijsbers2024amlb} and \citet{erickson2025tabarena}, we set the budget to 4 hours per dataset (see Appendix~\ref{app:automl-details} for additional details on the settings). AutoKeras and AutoPyTorch search over libraries of human-implemented neural components and are the natural baselines for our agentic MLP$^\dagger$. They lag behind it in both the single-model and the ensemble regime. One factor is that their libraries leave most of the pipeline fixed: neither modularizes numerical embeddings, target preprocessing, or inference, and categorical embedding has a single component in AutoPyTorch and none in AutoKeras. Their $44$ and $4$ interchangeable components sit mostly on preprocessing and the optimizer, against $57$--$61$ spread over all axes for the agentic MLP$^\dagger$ space (Appendix~\ref{app:automl-details}). For AutoGluon we use the \texttt{best\_quality} preset and pass it both the train and validation sets. AutoGluon fits a meta-learned portfolio of $110$ fixed configurations across seven conventional model families (LightGBM, CatBoost, XGBoost, random forest, extra trees, and two MLP variants; Appendix~\ref{app:automl-details}), $8$-fold bags each one, feeds the out-of-fold predictions to a second stack layer of the same families, and greedily weights the final predictions. This design is complementary to our approach, and we test the combination directly: AutoGluon (+A) receives 8 randomly-sampled configurations of our agentic RealMLP and TabM$^\dagger$ models. The augmented system outperforms the base. AutoGluon selects the agent-implemented models into its final weighted ensemble on $43$ of the $45$ datasets, assigning them $54\%$ of the ensemble weight on average.

\textbf{TabArena.} Finally, the conclusions transfer to the recent TabArena benchmark (51 datasets, official protocol; see Appendix~\ref{app:tabarena}). With tuning and ensembling, the agentic spaces improve the official Elo scores of MLP$^\dagger$, TabM, RealMLP, and TabICLv2, placing both the agentic TabICLv2 and the agentic RealMLP above AutoGluon with the same \texttt{best\_quality} portfolio of conventional models as above. The agentic RealMLP thus overtakes AutoGluon on TabArena but not in Figure~\ref{fig:main_results}. The likely reason is the protocol. Under TabArena, every configuration is itself an 8-fold cross-validated ensemble, i.e., the same bagging machinery that AutoGluon uses internally. Its greedy ensemble draws on all $200$ tuning configurations, which is possible since they are randomly sampled. In our benchmark, each configuration is fit once on the training subset, scored once on the validation subset, and yields a single set of test predictions. Our greedy ensemble draws on $100$ fresh random configurations ($20$ on large datasets), because the TPE trials of the single-model study are highly correlated and cannot be reused. Both choices are deliberate: TPE strengthens the individual models, and forgoing cross-validation admits larger datasets as well as datasets with grouped and temporal splits, for which cross-validation is not well defined. Figure~\ref{fig:main_results} therefore gives a more conservative view of the individual model families than TabArena does.

\begin{table}[t]
\caption{Greedy-ensemble diversity aggregated over all 45 datasets. $\rho$ is the mean pairwise test-prediction correlation among the selected members (lower $\rightarrow$ more diverse pool); $\Delta_{\mathrm{base}}$ is the mean relative improvement of the agentic space over the base space on the dataset-native metrics, reported for the pool's strongest single member and for the greedy ensemble built from the same pool.}
\label{tab:ensemble_summary}
\scriptsize
\setlength{\tabcolsep}{6pt}
\begin{center}
\begin{tabular}{ll cccc}
\toprule
 & & MLP$^\dagger$ & TabM$^\dagger$ & LightGBM & RealMLP \\
\midrule
\multirow{2}{*}{Member corr.\ $\rho$ $\downarrow$} & Base & $0.964$ & $0.981$ & $0.913$ & $0.974$ \\
 & Agentic & $0.892$ & $0.872$ & $0.799$ & $0.957$ \\
\midrule
\multirow{2}{*}{$\Delta_{\mathrm{base}}$ (\%) $\uparrow$} & Best member & $+0.1$ & $+0.4$ & $-1.0$ & $0.0$ \\
 & Ensemble & $+1.3$ & $+1.1$ & $+0.2$ & $+0.3$ \\
\bottomrule
\end{tabular}
\end{center}
\end{table}

\subsection{Agent cost and amortization}
\label{sec:agent-cost}

\begin{table}[t]
\caption{Agent resource usage per model family. The columns report output tokens and wall-clock generation time for the main agent (Claude Code) and the ablation agent (Codex); the agent is run once per model family, and token counts are reported instead of dollar cost because provider pricing changes over time.}
\label{tab:agent_cost}
\scriptsize
\setlength{\tabcolsep}{5.0pt}
\begin{center}
\begin{tabular}{lrrrr}
\toprule
\multirow{2}{*}{Method}
& \multicolumn{2}{c}{Agent output tokens}
& \multicolumn{2}{c}{Agent time} \\
\cmidrule(lr){2-3}
\cmidrule(lr){4-5}
& Claude & Codex & Claude & Codex \\
\midrule
MLP$^\dagger$      & 80,965 & 92,254 & 01:34:12 & 00:54:47 \\
RealMLP  & 90,579 & 102,542 & 01:24:12 & 01:10:47 \\
TabM$^\dagger$     & 64,142 & 71,865 & 01:40:09 & 01:04:22 \\
LightGBM & 46,221 & 36,961 & 00:43:09 & 00:34:22 \\
TabICLv2 & 36,221 & 32,456 & 00:30:09 & 00:24:02 \\
\bottomrule
\end{tabular}
\end{center}
\end{table}

In Table~\ref{tab:agent_cost}, we report the agent's output token usage (LLM thinking and generation) and wall-clock generation time for Claude Code Opus 4.8 and Codex GPT 5.5. We omit input and cached tokens, which are dominated by the fixed system prompt, tool signatures, and environment setup shipped with the agent. Given current LLM costs per token \citep{anthropic_claude_pricing_2026, openai_api_pricing_2026}, we estimate that a single hypothesis generation loop costs approximately $10$ USD. The agent is run once per model family, while HPO is run per dataset. This distinction is important: agent generation takes roughly $35$--$100$ minutes depending on model family and agent, but the cost is amortized across all later datasets unlike the autoresearch \citep{karpathy2026autoresearch}. The method is therefore most attractive when the same generated space will be reused across many datasets.

\section{Analysis}
\label{sec:analysis}

\subsection{Which modules matter?}
\label{sec:module-importance}

\begin{figure}[t]
    \centering
     \includegraphics[width=\linewidth]{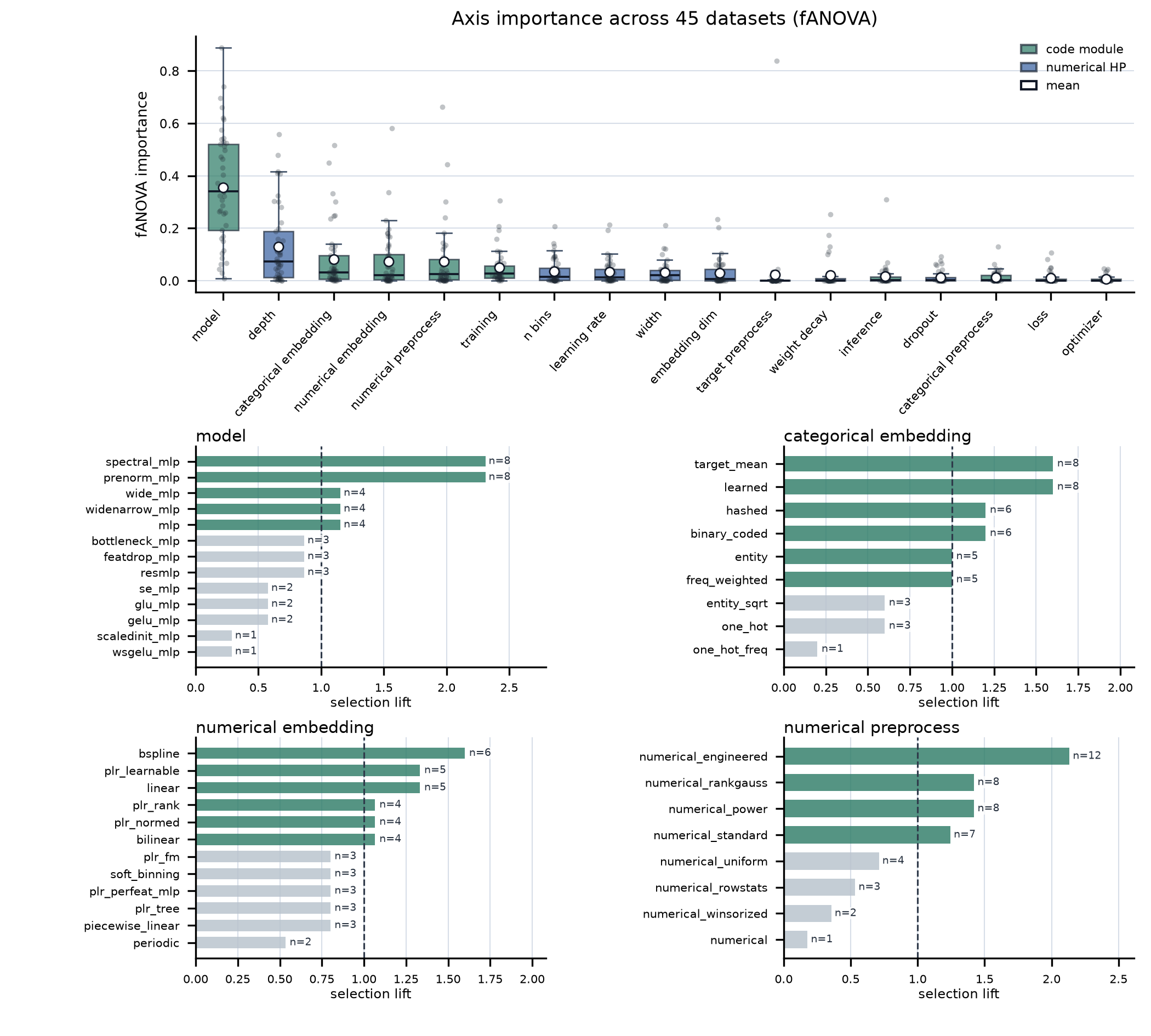}
     \caption{\emph{Top:} Module importance from fANOVA for MLP, normalized to sum to one per dataset. Green boxes are the code modules with agent-generated candidate implementations; blue boxes are conventional numerical hyperparameters. \emph{Bottom:} Selection lift of the candidate implementations of the four most important modules. For each candidate, $n$ is the number of datasets (out of 45) on which it appears in the best HPO trial, and the lift is $n$ divided by the count expected if the winner were drawn uniformly among the module's candidates; bars above one thus mark candidates selected more often than the uniform baseline.}
     \label{fig:module_lift}
\end{figure}

From the fANOVA chart of Figure~\ref{fig:module_lift}, we see substantial importance is assigned to several modules extended by the agent: numerical embeddings, architecture, training, and numerical preprocessing. Further, the per-candidate selection-lift plot of Figure~\ref{fig:module_lift} suggests that no single candidate dominates: the selections are spread across each module's candidate pool with no clean winner.

Aggregating the winning HPO trials across the 18 MLP regression datasets, we observe that the selected candidate combinations are pairwise distinct, and every candidate except two training-side ones appears in at least one of them. This suggests that the joint selection is necessary. Which candidates win is thus largely dataset-specific: the agent contributes a reusable basis of implementations, and the per-dataset composition has to be found by HPO. The candidate-pool ablation in Section~\ref{sec:candidate-pool} is consistent with this picture: no subsampled pool recovers the full space's performance.

\begin{table}[t]
\caption{Stability of agent-generated MLP$^\dagger$ search spaces across five independent re-generations per agent, evaluated on all 45 datasets. Mean rank, normalized score, and Elo (anchored at the base MLP$^\dagger$ = 1000) are computed in the joint pool of the 12 spaces (base, released, and the ten re-generations); improvement over MLP$^\dagger$ is the mean relative improvement on the dataset-native metrics. Agent rows report the mean $\pm$ half-width of a $95\%$ $t$-interval over the five re-generations; per-run values are in Table~\ref{tab:agent-full-ablation}.}
\label{tab:agent_ablation_summary}
\scriptsize
\setlength{\tabcolsep}{6pt}
\begin{center}
\begin{tabular}{l cccc}
\toprule
Search space & Mean rank $\downarrow$ & Norm.\ score $\uparrow$ & Elo $\uparrow$ & Improv.\ over MLP$^\dagger$ (\%) $\uparrow$ \\
\midrule
MLP$^\dagger$ (base space) & $8.49$ & $0.13$ & $1000$ & $0.0$ \\
Agentic MLP$^\dagger$ (released) & $6.96$ & $0.25$ & $1095$ & $+1.28$ \\
\midrule
Claude re-generations & $6.33 \pm 1.03$ & $0.25 \pm 0.09$ & $1132 \pm 61$ & $+0.78 \pm 0.63$ \\
Codex re-generations & $6.18 \pm 0.36$ & $0.26 \pm 0.06$ & $1141 \pm 22$ & $+0.97 \pm 0.42$ \\
\bottomrule
\end{tabular}
\end{center}
\end{table}

\subsection{Agent choice and stability}
\label{sec:agent-stability}

Table~\ref{tab:agent_ablation_summary} compares agent-generated MLP$^\dagger$ search spaces across five independent re-generations for Claude Code with Opus 4.8 at \texttt{max} thinking level and Codex with GPT-5.5 at \texttt{xhigh} thinking level \citep{anthropic2026opus48, anthropic2026claudecode, openai2025codex, openai2026gpt55}, evaluated on all 45 datasets. Every re-generated space improves over the base MLP$^\dagger$ space on all four aggregate metrics, and the released space used in the main experiments lies within the re-generation spread. The two agents perform on par: Codex is somewhat more consistent (Elo $1141 \pm 22$ vs.\ $1132 \pm 61$), while the best and the worst single runs both come from Claude (Elo $1182$ and $1056$). Per-run results are given in Table~\ref{tab:agent-full-ablation} in Appendix~\ref{app:ablations}.

\subsection{Candidate-pool size}
\label{sec:candidate-pool}

In Section~\ref{sec:module-importance}, we observe that the selected candidate combinations are pairwise distinct across datasets, which suggests that growing the search space should help. However, beyond some size, a larger search space could exacerbate overfitting to the validation subset. We check both claims by re-fitting the agentic MLP$^\dagger$ over progressively subsampled candidate pools: for each size in $\{8, 16, 32, 48\}$ we draw five random subsets of the candidates from the full pool ($61$ for regression, $57$ for classification), rebuild the search space, and re-run the full 200-trial HPO on 18 randomly chosen datasets. In Table~\ref{tab:module_subsample}, all four aggregate metrics improve monotonically with the pool size, and the full pool remains clearly the best (Elo $1143$ vs.\ $1086$ for the largest subsets): growing the search space indeed helps. We also observe no downturn at the current pool sizes, so overfitting does not yet set in at this scale.

\begin{table}[t]
\caption{Candidate-pool subsampling for the agentic MLP$^\dagger$ space on 18 randomly chosen datasets. For each pool size we draw five random subsets of the agent-generated candidate implementations, rebuild the search space, and re-run the full 200-trial HPO; the last row uses the complete pool. Metrics follow the main-figure definitions and are computed per replicate in a pool of six spaces (base, one random subset per size, full), with Elo anchored at the base MLP$^\dagger$ = 1000; each value is the mean $\pm$ half-width of a $95\%$ $t$-interval over the five replicates, so the intervals measure the sensitivity to which candidates are drawn. Base and full are the same space in every replicate; their small intervals reflect only the changing comparison pool.}
\label{tab:module_subsample}
\scriptsize
\setlength{\tabcolsep}{6pt}
\begin{center}
\begin{tabular}{l cccc}
\toprule
\# Candidates & Mean rank $\downarrow$ & Norm.\ score $\uparrow$ & Elo $\uparrow$ & Improv.\ over MLP$^\dagger$ (\%) $\uparrow$ \\
\midrule
0 (base) & $3.84 \pm 0.11$ & $0.18 \pm 0.05$ & $1000$ (anchor) & $0.0$ \\
8 & $3.52 \pm 0.42$ & $0.21 \pm 0.10$ & $1015 \pm 53$ & $+0.36 \pm 0.71$ \\
16 & $3.40 \pm 0.30$ & $0.24 \pm 0.14$ & $1026 \pm 40$ & $+0.49 \pm 0.59$ \\
32 & $3.33 \pm 0.55$ & $0.31 \pm 0.14$ & $1051 \pm 71$ & $+0.57 \pm 1.00$ \\
48 & $2.94 \pm 0.82$ & $0.35 \pm 0.20$ & $1086 \pm 105$ & $+0.73 \pm 0.56$ \\
Full & $2.68 \pm 0.18$ & $0.49 \pm 0.04$ & $1143 \pm 11$ & $+1.15$ \\
\bottomrule
\end{tabular}
\end{center}
\end{table}

\subsection{Trial budget.}
\begin{figure}[t]
    \centering
    \includegraphics[width=\linewidth]{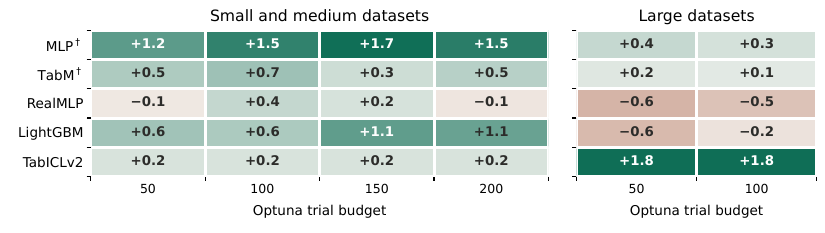}
    \caption{Effect of the matched Optuna trial budget for small-and-medium (left) and large (right) datasets. The cell color and value give the mean relative improvement of the agentic space over the corresponding base space at matched budget, computed on the dataset-native metrics (\%); green marks budgets where the agentic space is ahead. For TabICLv2, the cells do not vary with the budget, because the default TabICLv2 is not tuned and the agentic TabICLv2 is tuned at a fixed $100$ trials on every dataset.}
    \label{fig:budget_curve_main}
\end{figure}

The fixed trial-budget comparison of Figure~\ref{fig:main_results} could in principle depend on the chosen trial budget: a larger space needs more trials to be explored, so agentic spaces might only win once the budget is large enough. Figure~\ref{fig:budget_curve_main} therefore reports the average improvement of the agentic space over the base space at each matched Optuna budget. On small-and-medium datasets the gap is positive at every budget for MLP$^\dagger$, TabM$^\dagger$, and LightGBM, while the agentic RealMLP stays within $\pm0.4\%$ of its base space. On large datasets the gaps are smaller and mixed, with the exception of TabICLv2 ($+1.8\%$). The gaps change little from $50$ trials onwards, so the conclusions of Section~\ref{sec:main-performance} are not an artifact of a particular budget value.

\section{Limitations}
\label{sec:limitations}

The method expands a finite-budget HPO problem, so the superset relation $\Scal_{\mathrm{HPO}}\subseteq\Scal_{\mathrm{agent}}$ does not guarantee finite-budget improvement. Larger spaces can overfit validation. The agentic spaces can also raise the per-fit cost, since the agent may propose more expensive optimizers or test-time augmentations: the median fit-time ratio to the base space ranged from $0.9\times$ (RealMLP) to $2.4\times$ (TabICLv2) in our runs. Agent-generated code may potentially contain bugs; hence, it must be manually inspected before shipping. A potential improvement to the evaluation and HPO pipeline is to use cross-validation; however, this can be prohibitively expensive in practice. The approach also depends on the quality and reproducibility of the agent-generated candidate sets. Agent products, prompts, and retrieval corpora change over time, so they must be logged as experimental artifacts. We evaluate five model families and 45 datasets, which is broad enough to show the main pattern but not enough to claim universal gains across all tabular regimes. Our ensembling study covers four model families and reports a single greedy-ensembling run per (dataset, search space) over random candidate pools; extending it to foundation models, and to pools drawn from actual HPO trajectories, is left for future work. Finally, we intentionally avoid dataset-specific prompting during generation. This protects against leakage and amortizes cost, but it can miss modules that would be useful for a particular dataset's semantics or feature types. We detail some of the future research directions in Appendix~\ref{sec:research-directions}.

\section{Conclusion}
\label{sec:conclusion}

We propose to use LLM agents to generate structured, executable module implementations that expand the search space for HPO. By pairing this generated basis with a classical HPO algorithm, we obtain consistent gains for classic DNNs, while for GBDT, foundational, and heavily meta-tuned models the gains are smaller and concentrated in the regimes where the base model is not already saturated. The module and combination analyses suggest that the value comes from dataset-specific composition: no single generated candidate is universally best, but the candidate basis gives HPO more useful ways to adapt the model pipeline. This makes agentic search-space generation a practical middle ground between fixed HPO grids and fully autonomous research loops. We also find that agent-produced blocks add diversity that boosts ensemble performance: the greedy ensemble over the agentic RealMLP space is the strongest of all model families in our evaluation, ahead of a tuned tabular foundation model and behind only the multi-model AutoGluon system.

\bibliographystyle{unsrtnat}
\bibliography{references}

\newpage
\appendix

\section{Additional Results on TabArena}
\label{app:tabarena}

\textbf{Setup.} TabArena~\citep{erickson2025tabarena} evaluates each model family under a fixed protocol: one default configuration plus $200$ random configurations drawn from the family's search space, each fitted as an 8-fold bagged ensemble. \emph{Tuned} (T) reports the configuration with the best validation score, and \emph{Tuned+Ensembled} (T+E) reports the greedy weighted ensemble~\citep{caruana2004ensemble} built from the same $200$ configurations. We plug the agent-generated search spaces of Section~\ref{sec:proposed} into this protocol unchanged for the five families: LightGBM, MLP$^\dagger$, RealMLP, TabM, and TabICLv2. We mark the agentic version with a suffix \texttt{-A}. For the four families other than MLP$^\dagger$ the base numbers are the official TabArena entries; TabArena has no MLP entry, so we additionally run the base MLP$^\dagger$ ourselves under the identical protocol. Since the agentic design changes the tuning space and not the default configuration, MLP$^\dagger$ and MLP$^\dagger$-A share their default run, and only their Tuned and T+E results differ. Two differences from the setup of Section~\ref{sec:results} matter when comparing the numbers. First, TabArena tunes by random search over the $200$ configurations, whereas our benchmark obtains the tuned results with Optuna TPE, which is a stronger optimizer. Second, every TabArena configuration is itself an 8-fold cross-validated ensemble: it trains eight models, each on seven of the eight folds, scores the configuration on their out-of-fold predictions, and averages the test predictions of all eight models. Our benchmark trains each configuration once and scores it on the validation subset. Third, we evaluate on TabArena-Lite, i.e., the first of the seven outer cross-validation splits.

\begin{figure}[h]
    \centering
    \includegraphics[width=\linewidth]{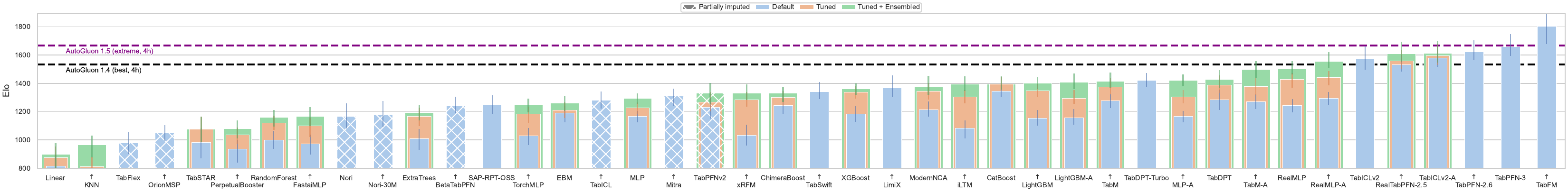}
    \caption{Official TabArena-Lite Elo leaderboard (51 datasets) with the agent-augmented spaces included as LightGBM-A, MLP-A, RealMLP-A, TabM-A, and TabICLv2-A. For each method the bars show the Elo of the default configuration, the best of $200$ random configurations (Tuned), and the greedy ensemble built from the same configurations (Tuned+Ensembled); whiskers are $95\%$ confidence intervals, dashed lines mark the AutoGluon reference systems, and hatched bars are partially imputed methods.}
    \label{fig:tabarena_elo}
\end{figure}

\textbf{Results.} Figure~\ref{fig:tabarena_elo} shows the official Elo leaderboard over the 51 TabArena-Lite datasets with the agentic variants included. The picture from Section~\ref{sec:main-performance} transfers: at the T+E level, the agentic space improves the Elo of MLP$^\dagger$ (from 1295 to 1421), TabM (from 1414 to 1497), RealMLP (from 1502 to 1557), and TabICLv2 (from 1573 to 1612), and roughly matches it for LightGBM (from 1402 to 1409). The ordering across families also matches Section~\ref{sec:main-performance}: the plain MLP$^\dagger$, whose base space is the least tuned, gains the most ($+125$), while the meta-tuned RealMLP and the saturated LightGBM gain the least. Notably, both TabICLv2-A and RealMLP-A at the T+E level outperform the \texttt{best\_quality} preset of AutoGluon~1.4 (1535)\footnote{Ensemble of LightGBM, CatBoost, XGBoost, random forests, and MLP.}, and TabICLv2-A is behind only the \texttt{extreme\_quality} preset of AutoGluon~1.5 (1669)\footnote{Adds TabPFNv2, TabICL, Mitra, TabM, RealTabPFN-2/2.5, TabDPT, TabPrep-LightGBM, and EBM to the 1.4 ensemble.} and newer TFMs. At the tuned-only level the gains are smaller ($+79$ for MLP$^\dagger$, $+14$ for RealMLP, $+3$ for TabM) or negative ($-53$ for LightGBM): random search is less sample-efficient in the larger agentic spaces, so, as in Section~\ref{sec:main-performance}, most of the benefit is realized when the diversity of the pool is exploited by ensembling. We note that Elo is fitted over the whole method pool, so all ratings shift slightly whenever the pool changes; the numbers above are therefore comparable within this leaderboard but not against other versions of it. Per-dataset Default/Tuned/T+E scores for all five families are reported in Appendix~\ref{app:tabarena-per-dataset}.

\section{Setup}
\label{app:setup-details}

\subsection{Hardware}
\label{app:setup-details-hardware}

Most of the experiments were conducted on a single NVIDIA A100 GPU paired with AMD EPYC CPU. Absolute wall-clock time depends on hardware, parallelism, and cluster load; the agent generation times of Table~\ref{tab:agent_cost} additionally depend on provider load and network latency, so they should be read as indicative rather than reproducible.

\subsection{Datasets}
\label{app:setup-details-data}

We provide a full list of datasets and their properties in Table~\ref{tab:per_dataset_statistics}, which also includes the evaluation metric and feature counts. The evaluation metric of each dataset is the native metric adopted from the prior work that introduced the corresponding benchmark (TabM~\citep{gorishniy2025tabm}, TabArena~\citep{erickson2025tabarena}, and TabReD~\citep{rubachev2024tabred}). The main text uses the aggregate summary in Table~\ref{tab:dataset_statistics}. We categorize datasets by size as follows: the \emph{large} group consists of the TabReD benchmark datasets together with Microsoft; the remaining datasets are split into \emph{small} and \emph{medium} at the median number of cells (rows $\times$ features), with the split boundary at $3.0\times10^{5}$ cells.

\begin{table}[h]
\caption{Per-dataset statistics. Size is the small/medium/large bucket used in the paper comparisons (large = TabReD datasets and Microsoft; the rest split at the median cell count). Metric is the dataset-native evaluation metric adopted from the source benchmark. \#Samples is the total number of rows. For regression we report \#Classes as 0.}
\label{tab:per_dataset_statistics}
\scriptsize
\setlength{\tabcolsep}{4pt}
\begin{center}
\begin{tabular}{llllrrrrrl}
\toprule
Dataset & Task & Size & Metric & \#Num. & \#Cat. & \#Bin. & \#Classes & \#Samples & Source \\
\midrule
airfoil\_self\_noise & reg & small & rmse & 4 & 1 & 0 & 0 & 1,503 & OpenML \\
Another-Dataset-on-used-Fiat-500 & reg & small & rmse & 6 & 1 & 0 & 0 & 1,538 & OpenML \\
black-friday & reg & medium & rmse & 5 & 4 & 0 & 0 & 166,821 & OpenML \\
california & reg & small & rmse & 8 & 0 & 0 & 0 & 20,640 & OpenML \\
concrete\_compressive\_strength & reg & small & rmse & 8 & 0 & 0 & 0 & 1,030 & OpenML \\
diamond & reg & medium & rmse & 6 & 3 & 0 & 0 & 53,940 & OpenML \\
Food\_Delivery\_Time & reg & medium & rmse & 6 & 3 & 0 & 0 & 45,451 & OpenML \\
healthcare\_insurance\_expenses & reg & small & rmse & 3 & 3 & 0 & 0 & 1,338 & OpenML \\
house & reg & medium & rmse & 16 & 0 & 0 & 0 & 22,784 & OpenML \\
miami\_housing & reg & small & rmse & 14 & 1 & 0 & 0 & 13,776 & OpenML \\
microsoft & reg & large & rmse & 136 & 0 & 0 & 0 & 1,200,192 & OpenML \\
physiochemical\_protein & reg & medium & rmse & 9 & 0 & 0 & 0 & 45,730 & OpenML \\
QSAR-TID-11 & reg & medium & rmse & 1024 & 0 & 0 & 0 & 5,742 & OpenML \\
QSAR\_fish\_toxicity & reg & small & rmse & 6 & 0 & 0 & 0 & 907 & OpenML \\
wine\_quality & reg & small & rmse & 11 & 1 & 0 & 0 & 6,497 & OpenML \\
cooking-time & reg & large & rmse & 186 & 3 & 3 & 0 & 319,986 & TabReD \\
delivery-eta & reg & large & rmse & 221 & 1 & 1 & 0 & 416,451 & TabReD \\
maps-routing & reg & large & rmse & 984 & 2 & 0 & 0 & 340,981 & TabReD \\
sberbank-housing & reg & large & rmse & 365 & 10 & 17 & 0 & 28,321 & TabReD \\
weather & reg & large & rmse & 100 & 0 & 3 & 0 & 423,795 & TabReD \\
\midrule
Amazon\_employee\_access & clf & small & roc\_auc & 0 & 9 & 0 & 2 & 32,769 & OpenML \\
APSFailure & clf & medium & roc\_auc & 170 & 0 & 0 & 2 & 76,000 & OpenML \\
bank-marketing & clf & medium & roc\_auc & 5 & 8 & 0 & 2 & 45,211 & OpenML \\
Bioresponse & clf & medium & roc\_auc & 1776 & 0 & 0 & 2 & 3,751 & OpenML \\
churn & clf & small & accuracy & 10 & 1 & 0 & 2 & 10,000 & OpenML \\
credit\_card\_clients\_default & clf & medium & roc\_auc & 20 & 3 & 0 & 2 & 30,000 & OpenML \\
customer\_satisfaction\_in\_airline & clf & medium & roc\_auc & 5 & 16 & 0 & 2 & 129,880 & OpenML \\
diabetes & clf & small & roc\_auc & 8 & 0 & 0 & 2 & 768 & OpenML \\
Diabetes130US & clf & medium & roc\_auc & 8 & 39 & 0 & 2 & 71,518 & OpenML \\
E-CommereShippingData & clf & small & roc\_auc & 6 & 4 & 0 & 2 & 10,999 & OpenML \\
GiveMeSomeCredit & clf & medium & roc\_auc & 10 & 0 & 0 & 2 & 150,000 & OpenML \\
heloc & clf & small & roc\_auc & 23 & 0 & 0 & 2 & 10,459 & OpenML \\
HR\_Analytics\_Job\_Change\_of\_Data\_Scientists & clf & small & roc\_auc & 2 & 10 & 0 & 2 & 19,158 & OpenML \\
in\_vehicle\_coupon\_recommendation & clf & medium & roc\_auc & 2 & 22 & 0 & 2 & 12,684 & OpenML \\
jm1 & clf & small & roc\_auc & 21 & 0 & 0 & 2 & 10,885 & OpenML \\
kddcup09\_appetency & clf & medium & roc\_auc & 174 & 38 & 0 & 2 & 50,000 & OpenML \\
NATICUSdroid & clf & medium & roc\_auc & 0 & 86 & 0 & 2 & 7,491 & OpenML \\
online\_shoppers\_intention & clf & small & roc\_auc & 10 & 6 & 1 & 2 & 12,330 & OpenML \\
polish\_companies\_bankruptcy & clf & medium & roc\_auc & 64 & 0 & 0 & 2 & 5,910 & OpenML \\
qsar-biodeg & clf & small & roc\_auc & 36 & 5 & 0 & 2 & 1,054 & OpenML \\
splice & clf & small & log\_loss & 0 & 60 & 0 & 3 & 3,190 & OpenML \\
taiwanese\_bankruptcy\_prediction & clf & medium & roc\_auc & 94 & 0 & 0 & 2 & 6,819 & OpenML \\
ecom-offers & clf & large & roc\_auc & 113 & 0 & 6 & 2 & 160,057 & TabReD \\
homecredit-default & clf & large & roc\_auc & 612 & 82 & 2 & 2 & 381,664 & TabReD \\
homesite-insurance & clf & large & roc\_auc & 253 & 23 & 23 & 2 & 260,753 & TabReD \\
\bottomrule
\end{tabular}
\end{center}
\end{table}

\subsection{Metrics}
\label{app:setup-details-metrics}

Each dataset is evaluated with its native metric (Appendix~\ref{app:setup-details-data}). To compare methods within a dataset, we use the signed score $s_B(m;\Dcal)$: the metric value at budget $B$ for lower-is-better metrics (RMSE, log-loss) and its negation for higher-is-better ones (AUROC, accuracy), so that lower $s_B$ is always better. We report four aggregate metrics; as discussed in \citet{erickson2025tabarena}, they offer complementary views on the performance.

\begin{itemize}[leftmargin=2em,itemsep=2pt,topsep=2pt]
    \item \textbf{Mean rank.} Pool every (model family, search space) pair on a dataset, sort by $s_B$, and let $\mathrm{rank}(m,\Scal;\Dcal)$ be the position of $(m,\Scal)$ in the sorted ordering (rank $1$ = best). The mean rank averages this across datasets.

    \item \textbf{Relative improvement.} Effect of switching to method $m$ from a reference method $r$ (the tuned base-space MLP$^\dagger$, unless stated otherwise; Table~\ref{tab:grouped_improvement} uses the model's own base space as $r$):
    \begin{equation}
    \label{eq:relative-improvement}
    \Delta_B(m;\Dcal)
        =
        100\,
        \frac{
            s_B(r;\Dcal)
            -
            s_B(m;\Dcal)
        }{
            \left|
                s_B(r;\Dcal)
            \right|
        },
    \end{equation}
    averaged across datasets. Because $\Delta_B$ is defined on the signed native metric, its denominator is the metric magnitude itself (e.g.\ $|\mathrm{AUROC}|$ rather than $|1-\mathrm{AUROC}|$, which would inflate small absolute changes on near-saturated datasets).

    \item \textbf{Normalized score.} The only metric computed on the error scale: $\err = 1 - \Score$ for the two higher-is-better metrics in our suite, i.e.\ $\err = 1 - \mathrm{AUROC}$ and $\err = 1 - \mathrm{accuracy}$, while the lower-is-better RMSE and log-loss are used as is, $\err = \Score$. Per dataset, errors are linearly rescaled so that the best method maps to $1$, the median method maps to $0$, and anything worse than the median is truncated to $0$; higher is better \citep{salinas2023tabrepo}.

    \item \textbf{Elo rating.} Each per-dataset comparison between two methods is a logistic ``match'' (the method with the better signed score wins); we fit ratings by minimizing the Bradley--Terry log-loss with L-BFGS over all pairs and dataset/budget combinations following \citet{chiang2024chatbot}, then shift the resulting scale so that the MLP$^\dagger$ baseline anchors at $1000$. Larger Elo means stronger overall performance.
\end{itemize}
Together, rank and Elo capture the relative standing, while relative improvement and normalized score summarize the scale of the change.

\subsection{Hyperparameters}
\label{app:setup-details-hps}
We use model-specific HPO spaces that match strong published tabular baselines, then compare each baseline against its agent-augmented counterpart under the same split, metric, sampler, and trial budget.
\paragraph{MLP$^\dagger$.}
Our MLP baseline follows the feed-forward baseline used with TabM: a stack of fully connected blocks with tuned depth, width, dropout, learning rate, weight decay, and piecewise-linear numerical embeddings~\citep{gorishniy2025tabm,gorishniy2022embeddings}. The PLR embedding bins and embedding dimension are part of the HPO space because numerical embeddings are a major driver of neural tabular performance. The full tuning space is shown in Table~\ref{tab:mlp_hpo_space}.

\begin{table}[h]
\caption{The hyperparameter tuning space for MLP$^\dagger$.}
\label{tab:mlp_hpo_space}
\begin{center}
\begin{tabular}{ll}
    \toprule
    Parameter & Distribution \\
    \midrule
    \# blocks & UniformInt[1, 6] \\
    Block width & UniformInt[64, 1024], step=16 \\
    Dropout rate & \{0.0, Uniform[0.0, 0.5]\} \\
    \midrule
    Learning rate & LogUniform[3e-5, 1e-3] \\
    Weight decay & \{0, LogUniform[1e-4, 1e-1]\} \\
    Gradient clipping & 1.0 \\
    Max epochs & 256 \\
    Early-stopping patience & 16 \\
    \midrule
    \# PLR bins on OpenML & UniformInt[16, 128], step=4 \\
    \# PLR bins on TabRed, Microsoft & UniformInt[8, 32] \\
    PLR embedding dim & UniformInt[8, 128], step=4 \\
    \midrule
    \# Optuna iterations / ensemble size on small-medium & 200/100 \\
    \# Optuna iterations / ensemble size on large & 100/20 \\
    \bottomrule
\end{tabular}
\end{center}
\end{table}

\paragraph{TabM$^\dagger$.}
TabM is a parameter-efficient neural ensemble that shares most parameters across ensemble members while producing $k$ predictions per example~\citep{gorishniy2025tabm}. We use the mini TabM configuration with $k=32$, tune the backbone width, depth, dropout, PLR embedding parameters, learning rate, and weight decay, and keep the same early-stopping and gradient-clipping protocol as for MLP. The full tuning space is shown in Table~\ref{tab:tabm_hpo_space}.

\begin{table}[h]
\caption{The hyperparameter tuning space for TabM$^\dagger$.}
\label{tab:tabm_hpo_space}
\begin{center}
\begin{tabular}{ll}
    \toprule
    Parameter & Distribution \\
    \midrule
    \# blocks & UniformInt[2, 4] \\
    Block width & UniformInt[64, 1024] \\
    Dropout rate & Uniform[0.0, 0.5] \\
    Ensemble size $k$ & 32  \\
    \midrule
    \# PLR bins on OpenML & UniformInt[2, 128] \\
    \# PLR bins on TabRed, Microsoft & UniformInt[8, 32] \\
    PLR embedding dim on OpenML & UniformInt[8, 32], step=4 \\
    PLR embedding dim on TabRed, Microsoft & UniformInt[8, 24], step=4 \\
    \midrule
    Learning rate & LogUniform[3e-5, 1e-3] \\
    Weight decay & \{0, LogUniform[1e-4, 1e-1]\} \\
    Gradient clipping & 1.0 \\
    Max epochs & 256 \\
    Early-stopping patience & 16 \\
    \midrule
    \# Optuna iterations / ensemble size on small-medium & 200/100 \\
    \# Optuna iterations / ensemble size on large & 100/20 \\
    \bottomrule
\end{tabular}
\end{center}
\end{table}

\paragraph{RealMLP.}
RealMLP is included as a strong pre-tuned modular MLP baseline designed to be competitive without extensive manual retuning~\citep{holzmueller2024realmlp}. We tune the RealMLP choices exposed in its recommended configuration family, including front scaling, dropout, activation, hidden-size template, numerical embedding type, PLR scale, learning rate, weight decay, and label smoothing for classification. The full tuning space is shown in Table~\ref{tab:realmlp_hpo_space}.

\begin{table}[h]
  \caption{The hyperparameter tuning space for RealMLP.}
  \label{tab:realmlp_hpo_space}
\begin{center}
  \begin{tabular}{ll}
      \toprule
      Parameter & Distribution / value \\
      \midrule
      Large configuration flag & Categorical\{False, True\} \\
      Hidden layout & \texttt{rectangular} \\
      \# hidden layers & IntUniform[2, 4] \\
      Hidden width & Categorical\{256, 384, 512\} \\
      Activation & \texttt{mish} \\
      Dropout $p$ & Uniform[0.0, 0.5] \\
      Dropout schedule & \texttt{flat\_cos} \\
      \midrule
      Embedding size & Categorical\{4, 8, 16\} \\
      PLR $\sigma$ & LogUniform[$10^{-2}$, 50] \\
      PLR learning-rate factor & LogUniform[$5\cdot10^{-2}$, $3\cdot10^{-1}$] \\
      PLR hidden widths (for large) & Categorical\{8, 16, 32, 64\}x2 \\
      PLR hidden widths (for not large) & (16, 4) \\
      Max one-hot category size & $\lfloor \mathrm{LogUniform}[4, 33] \rfloor$ \\
      \midrule
      Learning rate & LogUniform[$2\cdot10^{-2}$, $3\cdot10^{-1}$] \\
      Weight decay & LogUniform[$10^{-3}$, $5\cdot10^{-2}$] \\
      Scale learning-rate factor & LogUniform[2, 10] \\
      First-layer learning-rate factor & LogUniform[0.3, 1.5] \\
      $1 -$ squared momentum & LogUniform[$5\cdot10^{-3}$, $5\cdot10^{-2}$] \\
      Use label smoothing & Categorical\{False, True\} \\
      Label smoothing $\epsilon$ & LogUniform[$5\cdot10^{-3}$, $10^{-1}$] \\
      Label smoothing schedule & \texttt{coslog4} \\
      \midrule
      Epochs (for large) & Categorical\{256, 512\} \\
      Epochs (for not large) & 256 \\
      Early stopping (for large) & (3, 40) = (mult., add.) \\
      Early stopping (for not large) & disabled \\
      Ensemble size & 8 \\
      Ensemble averaging before softmax & False \\
      \midrule
    \# Optuna iterations / ensemble size on small-medium & 200/100 \\
    \# Optuna iterations / ensemble size on large & 100/20 \\
      \bottomrule
  \end{tabular}
\end{center}
  \end{table}

\paragraph{LightGBM.}
LightGBM is a histogram-based gradient-boosted decision tree method~\citep{ke2017lightgbm}. We use the HPO ranges adopted by the TabR benchmark protocol~\citep{gorishniy2024tabr}, tuning feature and bagging fractions, learning rate, leaf count, L2 regularization, and the minimum Hessian mass per leaf. The number of estimators is fixed at $4000$ with early stopping after $200$ rounds. The full tuning space is shown in Table~\ref{tab:lightgbm_hpo_space}.

\begin{table}[h]
\caption{The hyperparameter tuning space for LightGBM.}
\label{tab:lightgbm_hpo_space}
\begin{center}
\begin{tabular}{ll}
    \toprule
    Parameter & Distribution \\
    \midrule
    feature\_fraction & Uniform[0.5, 1.0] \\
    learning\_rate & LogUniform[1e-3, 1.0] \\
    num\_leaves & UniformInt[4, 768] \\
    min\_sum\_hessian\_in\_leaf & LogUniform[1e-4, 100] \\
    bagging\_fraction & Uniform[0.5, 1.0] \\
    lambda\_l2 & \{0, LogUniform[0.1, 10.0]\} \\
    n\_estimators & 4000 (fixed) \\
    bagging\_freq & 1 \\
    early\_stopping\_rounds & 200 \\
    \midrule
    \# Optuna iterations / ensemble size on small-medium & 200/100 \\
    \# Optuna iterations / ensemble size on large & 100/20 \\
    \bottomrule
\end{tabular}
\end{center}
\end{table}

\paragraph{TabICLv2.}
TabICLv2 is a tabular foundation model for in-context learning on large tabular datasets~\citep{qu2026tabiclv2}. We keep the published \texttt{TabICLv2} checkpoint fixed and do not train or fine-tune the backbone. We fit the model by passing the concatenated train and validation portions of the dataset as the context: since TabICLv2 performs no gradient-based training, the validation split is not needed for early stopping and would otherwise go unused, whereas the other families consume it for early stopping and model selection. To avoid out-of-memory issues, we subsample the context to the minimum of $40,000$ samples or ${3,840,000}/{n_\text{features}}$.

\subsection{Hyperparameter optimization and ensembling}
\label{app:setup-details-hpo}

\paragraph{Hyperparameter optimization.}
We use the univariate TPE sampler in Optuna to tune individual model hyperparameters~\citep{bergstra2011algorithms,akiba2019optuna}. For small and medium datasets, we set the budget to 200 trials; for large ones (TabReD and Microsoft), to 100 trials. The exception is TabICLv2: the default model is evaluated as published, without any tuning, and the agentic TabICLv2 is tuned with a fixed budget of 100 trials on all datasets. In both cases, the first 20 trials use the random sampler to encourage exploration of the space. Empirically, we find the default univariate TPE sampler with the above budgets to work well, and we leave the search for better defaults to future work. For example, the number of random startup trials could be chosen adaptively based on the Coupon Collector's problem \citep{tang_tpe_startup_trials}.

\paragraph{Ensembles.} We build greedy ensembles following \citet{caruana2004ensemble}. We first sample 100 random configurations (20 for large datasets) from the given HPO space and train a model for each. We then iteratively grow the ensemble for a fixed budget of steps: at each step, we try adding each fitted model (selection is with replacement), average its predictions with those of the already selected models, and keep the candidate that yields the lowest validation error. Note that we do not use early stopping: the ensemble is grown for the full budget even if the validation error stops improving at some intermediate step.

\section{Future Research Directions}
\label{sec:research-directions}
Beyond the aggregate gains, the per-axis selection patterns in
Figure~\ref{fig:module_lift} reveal a structural pattern in how tabular
research progresses. On three of the four most important modules (the
model, the training procedure, and the numerical preprocessing),
the non-default agent-proposed implementations are selected most often. The fourth
module, numerical embeddings, is the only one whose default remains competitive, and it is also the one that has received by far the most attention in the recent tabular deep
learning~\citep{gorishniy2022embeddings,gorishniy2025tabm,holzmueller2024realmlp}. The other three modules, by comparison, remain underexplored: while isolated efforts exist -- for instance, recent work on optimizers for tabular deep learning~\citep{gorishniy2026benchmarking} -- much else is
overlooked. Therefore, we believe more investigation into the other pipeline modules is needed.

We attribute this pattern to two complementary factors. First, there is a long tail of plausible-but-forgotten methods on each underexplored axis, displaced by inertia rather than by evidence. Second, the breadth of the hypothesis pool available to an agent far exceeds that of a human researcher: an agent can inexpensively ``swarm'' the literature for
plausible candidates whenever the user requests it, with no commitment to championing any single one. On the other hand, researchers and machine learning practitioners tend to concentrate on a few well-established options. The combination -- a long tail of forgotten methods and a wide retrieval window -- is, in our view, the principal reason agent-augmented HPO improves over fixed search spaces precisely on the underexplored axes. We view a systematic
study of such overlooked axes -- now cheap to enumerate via agentic generation -- as a productive direction in itself.

\section{Agent Setup and Artifacts}
\label{app:llm-skills-and-guardrails}

Between the runs and before the first use, we sanitize the agent environment and memory to avoid potential leaks or biases by running the following procedure:
\begin{itemize}[leftmargin=2em]
    \item Remove memory and session information via
    \begin{itemize}[leftmargin=2em]
        \item \texttt{rm -rf ~/.claude/projects/<...>/memory}
        \item \texttt{rm -rf ~/.claude/projects/<...>/sessions}
    \end{itemize}
    \item Checkout a clean branch: \texttt{git checkout -b <semiagent-version-x>}
    \item Sanitize \texttt{git} history
    \begin{itemize}[leftmargin=2em]
        \item \texttt{git reflog expire --expire=now --all}
        \item \texttt{git gc --prune=now --quiet}
    \end{itemize}
\end{itemize}

\subsection{AutoML baselines}
\label{app:automl-details}

\paragraph{Protocol.} We run AutoKeras 1.1.0 (TensorFlow 2.15), AutoPyTorch 0.2.1, and AutoGluon 1.5.0 unmodified on the 45 datasets with our exact test splits. We pass our validation split directly into the system: AutoPyTorch via a custom holdout, AutoKeras via \texttt{validation\_data}). AutoGluon performs its own internal bagging, so we pass it concatenated train and validation subsets. We set the budget to 4 hours for each system following \citet{erickson2025tabarena, gijsbers2024amlb}. AutoPyTorch returns both the best single model and the greedy ensemble. AutoKeras returns the best model. AutoGluon returns a single predictor, which is a bagged and stacked ensemble.

\paragraph{Library size.} Table~\ref{tab:automl_library_size} maps the search spaces of the two neural AutoML systems onto our module axes and counts interchangeable implementations, excluding the default on each axis. AutoPyTorch's $22$ numerical-preprocessing components are $7$ scalers and the variance threshold plus $15$ feature-preprocessing transforms it inherits from auto-sklearn \citep{feurer2015autosklearn}; its $8$ architecture components are $4$ backbones and $4$ weight initializers, and its $7$ training components are $6$ learning-rate schedulers and MixUp. AutoKeras exposes normalization and batch normalization as on/off switches and two optimizer alternatives; the rest of its structured-data space is numeric (depth, width, dropout, learning rate).

\begin{table}[t]
\caption{Interchangeable components of the two neural AutoML libraries and of the agentic MLP$^\dagger$ space, mapped onto our module axes; counts exclude the default implementation on each axis. Axes dispatched by task are reported as regression/classification.}
\label{tab:automl_library_size}
\scriptsize
\setlength{\tabcolsep}{6pt}
\begin{center}
\begin{tabular}{l ccc}
\toprule
Module axis & AutoPyTorch 0.2.1 & AutoKeras 1.1.0 & Agentic MLP$^\dagger$ \\
\midrule
Numerical preprocessing & 22 & 1 & 7 \\
Categorical preprocessing & 2 & 0 & 3 \\
Target preprocessing & 0 & 0 & 4/0 \\
Numerical embedding & 0 & 0 & 11 \\
Categorical embedding & 1 & 0 & 8 \\
Architecture & 8 & 1 & 12 \\
Training & 7 & 0 & 11 \\
Optimizer & 4 & 2 & 1 \\
Loss & 0 & 0 & 2/1 \\
Inference & 0 & 0 & 2/3 \\
\midrule
Total & 44 & 4 & 61/57 \\
\bottomrule
\end{tabular}
\end{center}
\end{table}

\paragraph{AutoGluon portfolio.} Under \texttt{best\_quality}, AutoGluon fits a fixed, meta-learned zero-shot portfolio of 110 configurations (Table~\ref{tab:ag_portfolio}), bags each configuration with 8-fold cross-validation, feeds the out-of-fold predictions to a second stack layer of the same model families, and combines the resulting models with greedy weighted ensembling. In the \texttt{extreme\_quality} preset, AutoGluon replaces the portfolio of 110 with a portfolio of 28 configurations which also includes TabM and tabular foundation models (RealTabPFN-v2, TabDPT, TabICL, Mitra).

\begin{table}[h]
\caption{Composition of AutoGluon's zero-shot portfolios: \texttt{best\_quality} (v1.5.0, as resolved in our runs) and \texttt{extreme\_quality} (v1.5, portfolio \texttt{zeroshot\_2025\_12\_18\_gpu}). Entries are the number of configurations per model family.}
\label{tab:ag_portfolio}
\scriptsize
\begin{center}
\begin{tabular}{llrr}
\toprule
AutoGluon key & Model family & \texttt{best\_quality} & \texttt{extreme\_quality} \\
\midrule
\texttt{GBM} & LightGBM & 16 & 5 \\
\texttt{GBM\_PREP} & TabPrep-LightGBM & -- & 5 \\
\texttt{CAT} & CatBoost & 20 & 1 \\
\texttt{XGB} & XGBoost & 10 & -- \\
\texttt{RF} & Random forest & 10 & -- \\
\texttt{XT} & Extra trees & 10 & -- \\
\texttt{NN\_TORCH} & PyTorch MLP & 21 & -- \\
\texttt{FASTAI} & FastAI tabular NN & 23 & -- \\
\texttt{TABM} & TabM & -- & 5 \\
\texttt{REALTABPFN-V2} & RealTabPFN-v2 & -- & 5 \\
\texttt{TABDPT} & TabDPT & -- & 5 \\
\texttt{TABICL} & TabICL & -- & 1 \\
\texttt{MITRA} & Mitra & -- & 1 \\
\midrule
Total & & 110 & 28 \\
\bottomrule
\end{tabular}
\end{center}
\end{table}

\paragraph{Agentic hand-off (AutoGluon (+A)).} We add 8 randomly-sampled configurations of the agentic RealMLP and TabM$^\dagger$ models. The additional agentic configurations are interleaved with the stock portfolio and are bagged, stacked, and ensembled exactly like the stock members under the same four-hour budget ($126$ configurations in total).

\subsection{Hypothesis-generation workflow}
\label{app:llm-skills-and-guardrails-hypo-workflow}

\begin{figure}[h]
  \centering
  \begin{verbatim}
  bin/modules/
  +-- embedding/
  |   +-- __init__.py
  |   +-- one_hot.py
  +-- loss_reg/
      +-- __init__.py
      +-- huber.py
  \end{verbatim}
  \caption{Module registry package structure under \texttt{bin/modules}. Each subdirectory corresponds to one pipeline stage; individual \texttt{.py} files implement registered module choices for that axis.}
  \label{fig:module-tree}
  \end{figure}

\paragraph{Agent environment.} Each model family is handled by a single agent run. The agent is given the repository for that family: a small pipeline in which each module axis (preprocessing, embedding, model, training, optimizer, loss, and inference) is represented by a package whose registry maps integer indices to builder functions (Figure~\ref{fig:module-tree}). The agent is equipped with two main tools: \texttt{inspect\_pipeline.py}, which prints the registered modules across all axes, and \texttt{test.py}, which runs a synthetic-data smoke test for a chosen module combination.
\paragraph{Workflow.} After a preflight check that the environment is set up and the baseline runs, the agent proceeds in four stages: \emph{inspection}, \emph{hypothesis generation}, \emph{implementation}, and \emph{testing}.
During \emph{inspection}, the agent maps out the repository structure and the model pipeline. During \emph{hypothesis generation}, it proposes up to \texttt{-{}-num-hypotheses} candidate ideas, spread across the axes, and filters out those that do not meet our evidence bar: an idea must (a) have a clear mechanism by which it affects the model, (b) have support in the literature, (c) not replicate something already present in the base model, and (d) be benign by default, i.e., not harm performance when inactive.
The surviving hypotheses are then implemented one at a time in an implement-and-test loop. For each module, the agent: (1) implements it as a new self-contained \texttt{.py} file; (2) registers it in the package's \texttt{\_\_init\_\_.py}; (3) runs the smoke test (\texttt{uv run test.py -{}-<axis>-idx <idx>}); (4) reads any performance and efficiency warnings; and (5) debugs the module, or drops it if it degrades runtime or memory by more than 10x times. After all hypotheses are processed, the agent tests module \emph{interactions} by running \texttt{uv run test.py -{}-exhaustive} and debugs or discards any problematic combinations. The full candidate sets are provided in Tables~\ref{tab:agent_modules_mlp}--\ref{tab:agent_modules_tabicl}.

 \paragraph{Synthetic dataset test.} The \texttt{test.py} script evaluates each candidate against the all-\texttt{v0} baseline on 4 temporary synthetic datasets generated inside the test runner. Each scenario has \(N=N_{\mathrm{train}}+N_{\mathrm{val}}+N_{\mathrm{test}}\) rows, \(p\) numerical features,
  and \(q\) categorical features, with \((p,q)\in\{(100,100),(12,12),(16,0),(0,16)\}\). For numerical features, the runner samples
  \[
  x^{\mathrm{num}}_{ij}=z_{ij}+t_i a_j,\qquad
  z_{ij}\sim\mathcal{N}(0,1),
  \]
  where \(t_i\) is linearly spaced from \(-1\) to \(1\) over rows and \(a_j\) is linearly spaced from \(0.05\) to \(0.5\) over numerical
  columns. Categorical features are sampled as integer codes. By default,
  \[
  x^{\mathrm{cat}}_{ij}\sim \mathrm{Unif}\{0,\ldots,5\}.
  \]
  In the \((12,12)\) categorical edge-case scenario, the first four categorical columns use
  \[
  x^{\mathrm{cat}}_{ij}\sim \mathrm{Unif}\{0,\ldots,3\}\quad \text{on train},\qquad
  x^{\mathrm{cat}}_{ij}\sim \mathrm{Unif}\{0,\ldots,4\}\quad \text{on validation/test},
  \]
  with category \(4\) forced to appear outside training. The next four columns use the converse stress case,
  \[
  x^{\mathrm{cat}}_{ij}\sim \mathrm{Unif}\{0,\ldots,6\}\quad \text{on train},\qquad
  x^{\mathrm{cat}}_{ij}\sim \mathrm{Unif}\{0,\ldots,2\}\quad \text{on validation/test},
  \]
  with category \(6\) forced to appear in training. Remaining categorical columns use \(\mathrm{Unif}\{0,\ldots,4\}\) in all splits.

  Targets are generated from a finite signal plus noise. Let
  \[
  s_i
  =
  \sum_{j=1}^{\min(p,8)} b_j x^{\mathrm{num}}_{ij}
  +
  \sum_{j=1}^{\min(q,8)} c_j (x^{\mathrm{cat}}_{ij}\bmod 7)
  +
  \epsilon_i,
  \qquad
  \epsilon_i\sim\mathcal{N}(0,0.1^2),
  \]
  where \(b_j\) is linearly spaced from \(0.2\) to \(1.0\) and \(c_j\) from \(0.03\) to \(0.11\). Regression targets are \(y_i=s_i\).
  Binary labels are \(y_i=\mathbf{1}\{s_i>\mathrm{median}_{k\in\mathrm{train}}(s_k)\}\). Multiclass labels are obtained by discretizing
  \(s_i\) at the training-set tertiles. The runner uses a short two-epoch pass by default and a twelve-epoch long pass to expose modules that activate only after warmup or several epochs, such as schedulers, exponential moving averages~\citep{polyak1992acceleration}, or
  training-loop changes. It checks that predictions and metrics are finite, and reports wall time, iteration speed, trainable parameter
  count, process memory, and GPU memory when CUDA is used.

\subsection{Autoresearch details}
\label{app:autore-prog}

\begin{figure}[h]
    \centering
    \IfFileExists{resources/autoresearch_loop.pdf}{%
        \includegraphics[width=1\linewidth]{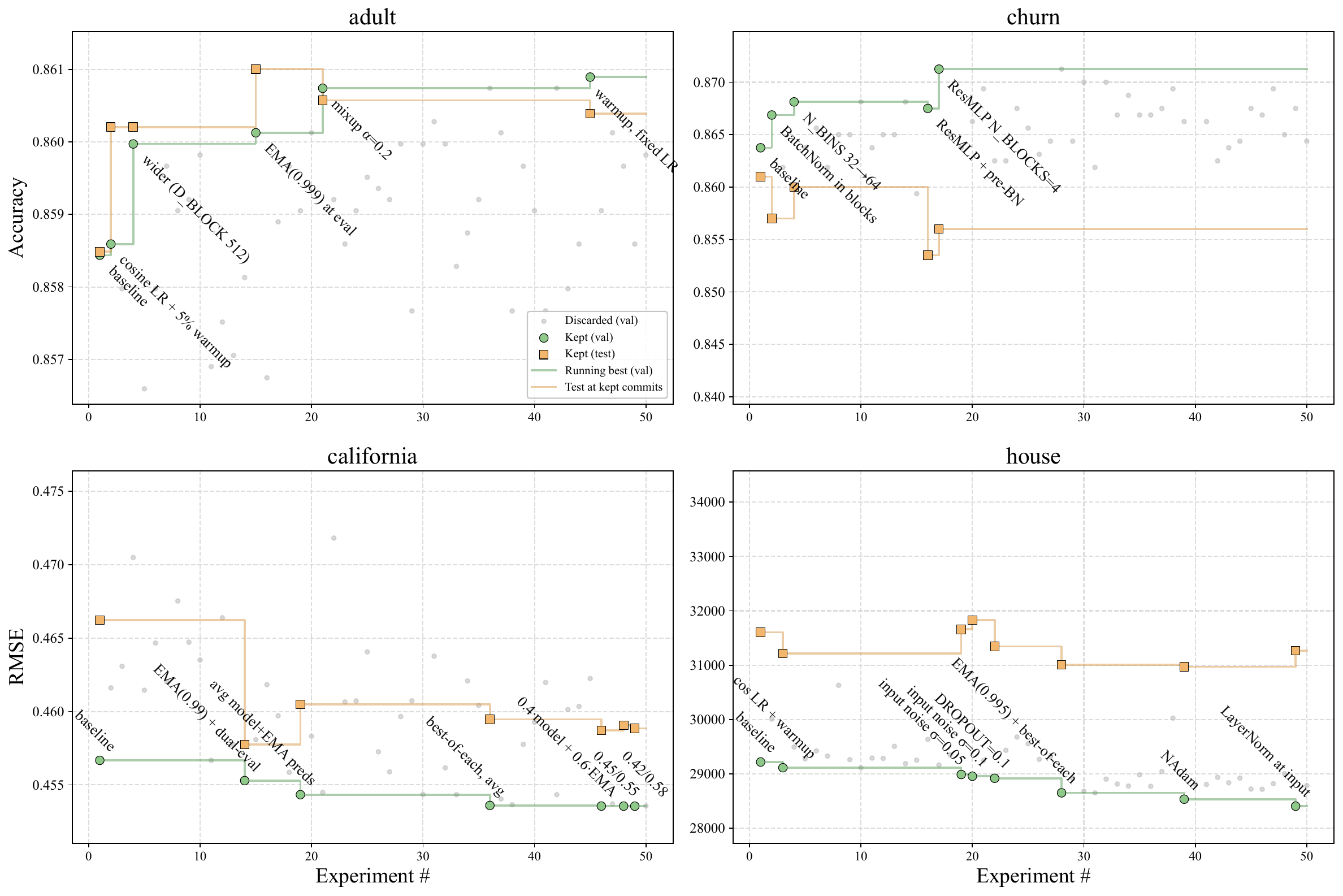}%
    }{%
        \placeholdergraphic[1\linewidth]{Autoresearch figure placeholder. Current runs: 50 iterations; best-so-far and final score remain near the fixed-HPO baseline rather than converging to a strong solution.}%
    }
    \caption{Autoresearch on $\mathrm{MLP}^\dagger$: 50 iterations on four
datasets. Validation improves only marginally and trails classical HPO and our agentic~HPO; test
does not consistently follow validation.}
    \label{fig:autoresearch}
\end{figure}

We adapt the autoresearch repository framework for tabular data and describe its mechanics below \citep{karpathy2026autoresearch}. Figure~\ref{fig:autoresearch} shows the per-iteration trajectories on the four evaluated datasets. In Table~\ref{tab:autoresearch_vs_hpo}, we compare the autoresearch best proposed solution to the matched trial-budget solutions obtained by the agentic and classical HPO. We see that the classical HPO matches or beats autoresearch on three of the four datasets, while agentic HPO further improves on the classical version everywhere. We note that due to the cost (varying from $20$ to $50$ USD per run), we do not re-run the autoresearch, so we report a single run without error bars.

\Needspace{6\baselineskip}

\begin{table}[h]
\caption{Comparison of Autoresearch, Default HPO and Agentic HPO at a budget
of 50 trials each on the $\mathrm{MLP}^\dagger$ pipeline.}
\label{tab:autoresearch_vs_hpo}
\scriptsize
\setlength{\tabcolsep}{6pt}
\renewcommand{\arraystretch}{1.10}
\begin{center}
\begin{tabular}{lrrrr}
\toprule
Method        & {adult $\uparrow$} & {churn $\uparrow$} & {california $\downarrow$} & {house $\downarrow$} \\
\midrule
Autoresearch  & $0.8604$              & $0.8560$              & $0.4589$              & $31{,}265$               \\
Default HPO   & $0.8626 \pm 0.0011$   & $0.8599 \pm 0.0018$   & $0.4584 \pm 0.0036$   & $31{,}361 \pm 356$  \\
Agentic HPO   & $0.8671 \pm 0.0011$   & $0.8621 \pm 0.0031$   & $0.4532 \pm 0.0030$   & $30{,}335 \pm 201$  \\
\bottomrule
\end{tabular}
\end{center}
\end{table}

Unlike our method, which uses the agent only to propose candidate module implementations, the autoresearch baseline of Section~\ref{sec:proposed} lets the agent edit the model end to end. It runs an autonomous loop on a single regression or classification task. Starting from the baseline \texttt{train.py}, each step (i) edits \texttt{train.py}---any hyperparameter, preprocessing, embedding, optimizer, loss, training-loop, or architecture change is allowed; (ii) commits the change; (iii) runs \texttt{benchmark.py}, which prints a single validation score; and (iv) keeps the commit if validation improves and reverts it (\texttt{git reset -{}-hard}) otherwise, appending every trial to a results log. The data loader \texttt{prepare.py} and the entry point \texttt{benchmark.py} are read-only, the held-out test split is never exposed, and the model must remain a single network of the corresponding family (no model swap, no ensembling of independently trained models, no new dependencies). The loop runs for a fixed iteration count. Because it edits one dataset at a time and accepts changes greedily on validation, it neither amortizes generation across datasets nor captures the module interactions that joint HPO exploits, which is reflected in its weaker results in Table~\ref{tab:autoresearch_vs_hpo}.


\Needspace{12\baselineskip}
\captionof{table}{Agent-proposed module candidates for MLP$^\dagger$ (67 candidates, deduplicated across the regression and classification branches). Counts exclude the default \texttt{v0} module on each axis.}
\label{tab:agent_modules_mlp}
\begin{center}
\scriptsize
\setlength{\tabcolsep}{4pt}
\renewcommand{\arraystretch}{1.15}
\begin{tabularx}{\textwidth}{@{}l >{\raggedright\arraybackslash}X@{}}
\toprule
Axis & Candidate variants \\
\midrule
Numerical preprocessing & Engineered features, power/Yeo--Johnson, RankGauss, row statistics, standardization, uniform-quantile, winsorization, trimmed-MAD \citep{yeo2000power} \\
Categorical preprocessing & Hashing, rare-bucketing, target-ordered encoding \citep{weinberger2009featurehashing,miccibarreca2001preprocessing} \\
Target preprocessing & Log, MAD, quantile, symlog target transforms (regression) \\
Numerical embedding & Linear, piecewise-linear, PLR (learnable, normed, tree, rank, per-feature MLP, Fourier-mapped), B-spline, periodic, soft-binning, bilinear \citep{gorishniy2022embeddings,rahimi2007randomfeatures} \\
Categorical embedding & One-hot, one-hot+frequency, learned/entity, sqrt-sized entity, frequency-weighted, binary-coded, hashed, target-mean \citep{guo2016entity,weinberger2009featurehashing} \\
Architecture & Bottleneck, feature-dropout, GELU, GLU, pre-norm, residual, scaled-init, squeeze--excitation, spectral-norm, wide, wide--narrow, weight-standardized GELU \citep{dauphin2017glu,shazeer2020glu,he2016resnet,hu2018squeeze} \\
Training & Cosine and warm-restart schedules, EMA, SWA, model soup, snapshot ensemble, gradient noise, MixUp, swap-noise, stochastic weight perturbation \citep{loshchilov2017sgdr,polyak1992acceleration,izmailov2018swa,zhang2018mixup} \\
Optimizer & Lookahead wrapper \citep{zhang2019lookahead} \\
Loss & Huber and Gaussian-NLL regression losses; label smoothing (classification) \citep{huber1964robust,szegedy2016label} \\
Inference & Train-range clipping, MC-dropout (regression); Gaussian-noise TTA, temperature scaling (classification) \citep{gal2016dropout,guo2017calibration} \\
\bottomrule
\end{tabularx}
\end{center}

\Needspace{12\baselineskip}
\captionof{table}{Agent-proposed module candidates for TabM$^\dagger$ \citep{gorishniy2025tabm} (56 candidates). Counts exclude the default \texttt{v0} module on each axis.}
\label{tab:agent_modules_tabm}
\begin{center}
\scriptsize
\setlength{\tabcolsep}{4pt}
\renewcommand{\arraystretch}{1.15}
\begin{tabularx}{\textwidth}{@{}l >{\raggedright\arraybackslash}X@{}}
\toprule
Axis & Candidate variants \\
\midrule
Numerical preprocessing & Clipped-quantile, missing indicators, log1p, robust/IQR, standardization, k-means-augmented quantile, rank-uniform, Yeo--Johnson \citep{yeo2000power,macqueen1967kmeans} \\
Categorical preprocessing & Frequency encoding, hashing \citep{weinberger2009featurehashing} \\
Target preprocessing & Log1p, quantile, robust target transforms (regression) \\
Numerical embedding & Advanced PLR, gated PLR, piecewise-linear, polynomial, tokenized \citep{gorishniy2022embeddings} \\
Categorical embedding & One-hot, one-hot+learned, one-hot with unknown bucket, learned, target-mean \citep{miccibarreca2001preprocessing} \\
Architecture & Bilinear, column/feature dropout, FiLM, GeGLU, SwiGLU+RMSNorm, input gates, linear-residual, LoRA, pre-norm, squeeze--excitation, soft-MoE, sparse-hidden, wide-shallow, categorical-embedded \citep{perez2018film,shazeer2020glu,zhang2019rmsnorm,hu2022lora,puigcerver2024softmoe} \\
Training & Cosine schedule, CutMix, EMA, MixUp, negative-correlation, SAM, SWA, head decorrelation \citep{polyak1992acceleration,zhang2018mixup,yun2019cutmix,foret2021sam,izmailov2018swa} \\
Optimizer & Lion \citep{chen2023lion} \\
Loss & Huber regression loss \citep{huber1964robust} \\
Inference & Clipped, geometric, median, trimmed-mean, quantile-average, variance-weighted head aggregation, Gaussian-noise TTA (regression); temperature scaling (classification) \citep{lakshminarayanan2017deepensembles,guo2017calibration} \\
\bottomrule
\end{tabularx}
\end{center}

\Needspace{12\baselineskip}
\captionof{table}{Agent-proposed module candidates for RealMLP \citep{holzmueller2024realmlp} (13 candidates). Counts exclude the default \texttt{v0} module on each axis.}
\label{tab:agent_modules_realmlp}
\begin{center}
\scriptsize
\setlength{\tabcolsep}{4pt}
\renewcommand{\arraystretch}{1.15}
\begin{tabularx}{\textwidth}{@{}l >{\raggedright\arraybackslash}X@{}}
\toprule
Axis & Candidate variants \\
\midrule
Numerical preprocessing & Normal-quantile, Yeo--Johnson power transform \citep{gorishniy2021revisiting,yeo2000power} \\
Categorical preprocessing & Rare-category grouping, log-frequency count encoding \\
Numerical embedding & Piecewise-linear embedding \citep{gorishniy2022embeddings} \\
Categorical embedding & Shared-vocabulary embeddings \citep{gorishniy2025tabm} \\
Architecture & BatchEnsemble layers, pre-activation residual network \citep{wen2020batchensemble,he2016resnet} \\
Training & Exponential moving-average (EMA) weights, MixUp augmentation \citep{polyak1992acceleration,zhang2018mixup} \\
Optimizer & Global-norm gradient clipping \citep{pascanu2013difficulty} \\
Loss & Focal loss (classification); Huber loss (regression) \citep{lin2017focal,huber1964robust} \\
\bottomrule
\end{tabularx}
\end{center}

\Needspace{12\baselineskip}
\captionof{table}{Agent-proposed module candidates for LightGBM \citep{ke2017lightgbm} (48 candidates). The base learner is fixed; candidates act before (feature, augmentation) or after (post-processing) the learner. Counts exclude the default \texttt{v0} module on each axis.}
\label{tab:agent_modules_lightgbm}
\begin{center}
\scriptsize
\setlength{\tabcolsep}{4pt}
\renewcommand{\arraystretch}{1.15}
\begin{tabularx}{\textwidth}{@{}l >{\raggedright\arraybackslash}X@{}}
\toprule
Axis & Candidate variants \\
\midrule
Feature engineering & Target (k-fold) / frequency / groupby-mean encoding, categorical pair-concat and target-std, high-cardinality hashing, k-means cluster/distance, PCA, random projection, pairwise products, pairwise differences, quantile-bin, rank-percentile, row statistics, log1p-skew, NaN indicators/row-count, rare-category grouping, extreme counts, top-$k$ squares \citep{miccibarreca2001preprocessing,macqueen1967kmeans} \\
Data augmentation & (Stratified) bootstrap, Gaussian noise, uniform/rank jitter, column-swap and intra-target-swap noise, categorical/feature dropout, mean-imputation dropout, MixUp, minority oversampling, class undersampling, random duplication, low-variance swap, target-quantile bootstrap, perturbed-copy concat, outlier trimming, within-class interpolation \citep{efron1979bootstrap,zhang2018mixup} \\
Post-processing & Train-range/quantile clipping and renormalization, IQR winsorization, soft-clip, integer rounding, non-negativity clip, probability floor/clip, power sharpen/smooth, temperature scaling (smoothed), prior correction, prior-blend, train-prior balancing, quantile matching, std-matching, mean shrinkage, snap-to-uniques, label smoothing \citep{guo2017calibration,szegedy2016label} \\
\bottomrule
\end{tabularx}
\end{center}

\Needspace{12\baselineskip}
  \captionof{table}{Agent-proposed module candidates and implementation baskets for TabICLv2 \citep{qu2025tabicl,qu2026tabiclv2} (standard module
  pool: 14). The foundation-model backbone is frozen, so candidates act on inputs (preprocessing, permutation, augmentation) or outputs
  (ensembling, post-processing). Basket rows indicate module baskets used for ensembles.}
  \label{tab:agent_modules_tabicl}
\begin{center}
  \scriptsize
  \setlength{\tabcolsep}{4pt}
  \renewcommand{\arraystretch}{1.15}
  \begin{tabularx}{\textwidth}{@{}l >{\raggedright\arraybackslash}X@{}}
  \toprule
  Axis & Candidate variants / basket choices \\
  \midrule
  Feature preprocessing & Power, quantile-normal, quantile-uniform, quantile (no standardization) \citep{gorishniy2021revisiting}
  \\
  Feature permutation & Latin-square, random, and shift feature permutations \citep{qu2025tabicl} \\
  Data augmentation & Feature-noise \citep{qu2026tabiclv2} \\
  Ensembling & Mean, median, geometric-mean, and
  uncertainty-weighted \citep{caruana2004ensemble,lakshminarayanan2017deepensembles} \\
  Post-processing &  log1p,
  standardize (regression) \citep{guo2017calibration} \\
  \addlinespace[2pt]
  \midrule
  Feature-preproc baskets & \textbf{0}: identity + power; \textbf{1}: identity + power +
  quantile-normal + quantile-uniform; \textbf{2}: identity + power + raw quantile-normal \\
  Augmentation baskets & \textbf{0}: identity / no augmentation; \textbf{1}: identity + feature-noise
  augmentation \\
  Postproc baskets (clf.) & \textbf{0}: probability readout; \textbf{1}: probability readout + temperature
  scaling; \textbf{2}: probability readout + prior-blend \\
  Postproc baskets (reg.) & \textbf{0}: standardize; \textbf{1}: standardize + log1p\\
  \bottomrule
  \end{tabularx}
\end{center}

\section{Extended Results}
\label{app:extended-results}

This appendix reports extended results. The default variant searches only the model-specific space $\Scal_{\mathrm{HPO}}$ described in Appendix~\ref{app:setup-details}; the agentic variant searches the augmented space $\Scal_{\mathrm{agent}}$, which contains the default module choices as a subset. We provide the full list of agent-generated hypotheses in Tables~\ref{tab:agent_modules_mlp}--\ref{tab:agent_modules_tabicl}.

For each model family, the agent-generated candidate set is produced once and then reused across all datasets. The agent does not receive dataset contents or dataset-specific feedback during generation. Downstream selection is performed only by HPO: each trial is trained on the training split, selected by validation performance, and evaluated on the held-out test split after selection. We report the mean and standard deviation of the test metric over repeated evaluation seeds.

\subsection{Ablation study}
\label{app:ablations}

\begin{table}[t]
\caption{Per-run results behind Table~\ref{tab:agent_ablation_summary}: each row is one agent-generated MLP$^\dagger$ search space evaluated on all 45 datasets in the joint 12-space pool. Win rate is the fraction of datasets on which the run improves over the tuned base-space MLP$^\dagger$.}
\label{tab:agent-full-ablation}
\scriptsize
\setlength{\tabcolsep}{6pt}
\begin{center}
\begin{tabular}{l ccccc}
\toprule
Run & Mean rank $\downarrow$ & Norm.\ score $\uparrow$ & Elo $\uparrow$ & Improv.\ over MLP$^\dagger$ (\%) $\uparrow$ & Win rate (\%) $\uparrow$ \\
\midrule
Released (Claude) & $6.96$ & $0.25$ & $1095$ & $+1.28$ & $60$ \\
\midrule
Claude run 1 & $6.64$ & $0.21$ & $1114$ & $+0.43$ & $64$ \\
Claude run 2 & $5.78$ & $0.28$ & $1165$ & $+0.33$ & $76$ \\
Claude run 5 & $6.16$ & $0.26$ & $1142$ & $+1.28$ & $69$ \\
Claude run 6 & $5.49$ & $0.35$ & $1182$ & $+1.39$ & $73$ \\
Claude run 7 & $7.60$ & $0.15$ & $1056$ & $+0.49$ & $60$ \\
\midrule
Codex run 0 & $5.73$ & $0.32$ & $1168$ & $+0.59$ & $73$ \\
Codex run 1 & $6.18$ & $0.29$ & $1141$ & $+0.62$ & $69$ \\
Codex run 2 & $6.56$ & $0.22$ & $1119$ & $+1.22$ & $67$ \\
Codex run 3 & $6.24$ & $0.24$ & $1137$ & $+1.16$ & $69$ \\
Codex run 4 & $6.18$ & $0.22$ & $1141$ & $+1.27$ & $69$ \\
\bottomrule
\end{tabular}
\end{center}
\end{table}

Table~\ref{tab:agent-full-ablation} reports the per-run agent results behind the summary in Section~\ref{sec:agent-stability}. All runs use the agentic MLP$^\dagger$ pipeline under the same split, metric, validation rule, and trial budget as the main experiments.

\subsection{Full per-dataset results}

We report full per-dataset results extracted at the terminal budget: 200 trials for small and medium datasets and 100 trials for the large ones in Table~\ref{tab:extended_per_dataset_results}. Below the family rows, each subtable also lists the individual-model rows of AutoPyTorch and AutoKeras; these are single unseeded runs, so no standard deviation is reported.

\Needspace{18\baselineskip}
\begin{center}
\captionof{table}{Individual results for each of the 45 datasets. Each subtable reports the terminal-budget mean test metric $\pm$ standard deviation for Default HPO and Agentic HPO under the same split, metric, and HPO budget. Arrows indicate whether higher or lower values are better.}
\label{tab:extended_per_dataset_results}
\end{center}

\Needspace{14\baselineskip}
\begin{center}
\scriptsize
\setlength{\tabcolsep}{4pt}
\renewcommand{\arraystretch}{1.04}

\begin{minipage}[t]{0.49\linewidth}
\centering
\textbf{black-friday ($\times 10^{3}$) $\downarrow$}\\
\begin{tabular}{lcc}
\toprule
Model & Default HPO & Agentic HPO \\
\midrule
MLP      & $3.4728 \pm 0.0025$ & $3.4779 \pm 0.0034$ \\
RealMLP  & $3.4497 \pm 0.0019$ & $3.4369 \pm 0.0010$ \\
LightGBM & $3.4516 \pm 0.0005$ & $3.4501 \pm 0.0004$ \\
TabM     & $3.4551 \pm 0.0032$ & $3.4484 \pm 0.0018$ \\
TabICL   & $3.5368 \pm 0.0034$ & $3.5344 \pm 0.0020$ \\
\midrule
AutoPyTorch & \multicolumn{2}{c}{$3.5426$} \\
AutoKeras   & \multicolumn{2}{c}{$3.6116$} \\
\bottomrule
\end{tabular}
\end{minipage}
\hfill
\begin{minipage}[t]{0.49\linewidth}
\centering
\textbf{california ($\times 10^{-1}$) $\downarrow$}\\
\begin{tabular}{lcc}
\toprule
Model & Default HPO & Agentic HPO \\
\midrule
MLP      & $4.5350 \pm 0.0199$ & $4.3717 \pm 0.0188$ \\
RealMLP  & $4.0618 \pm 0.0222$ & $4.0709 \pm 0.0179$ \\
LightGBM & $4.3319 \pm 0.0153$ & $4.3344 \pm 0.0172$ \\
TabM     & $4.3025 \pm 0.0228$ & $4.1363 \pm 0.0212$ \\
TabICL   & $3.9784 \pm 0.0062$ & $3.8974 \pm 0.0193$ \\
\midrule
AutoPyTorch & \multicolumn{2}{c}{$5.0470$} \\
AutoKeras   & \multicolumn{2}{c}{$5.0945$} \\
\bottomrule
\end{tabular}
\end{minipage}

\end{center}

\Needspace{14\baselineskip}
\begin{center}
\scriptsize
\setlength{\tabcolsep}{4pt}
\renewcommand{\arraystretch}{1.04}

\begin{minipage}[t]{0.49\linewidth}
\centering
\textbf{churn ($\times 10^{-1}$) $\uparrow$}\\
\begin{tabular}{lcc}
\toprule
Model & Default HPO & Agentic HPO \\
\midrule
MLP      & $8.6110 \pm 0.0275$ & $8.6003 \pm 0.0301$ \\
RealMLP  & $8.6223 \pm 0.0133$ & $8.6220 \pm 0.0171$ \\
LightGBM & $8.6033 \pm 0.0092$ & $8.5923 \pm 0.0216$ \\
TabM     & $8.6210 \pm 0.0124$ & $8.6143 \pm 0.0226$ \\
TabICL   & $8.6407 \pm 0.0147$ & $8.6500 \pm 0.0140$ \\
\midrule
AutoPyTorch & \multicolumn{2}{c}{$8.5500$} \\
AutoKeras   & \multicolumn{2}{c}{$8.5200$} \\
\bottomrule
\end{tabular}
\end{minipage}
\hfill
\begin{minipage}[t]{0.49\linewidth}
\centering
\textbf{diamond ($\times 10^{2}$) $\downarrow$}\\
\begin{tabular}{lcc}
\toprule
Model & Default HPO & Agentic HPO \\
\midrule
MLP      & $5.2571 \pm 0.0280$ & $5.2047 \pm 0.0321$ \\
RealMLP  & $5.2679 \pm 0.0199$ & $5.1885 \pm 0.0200$ \\
LightGBM & $5.3933 \pm 0.0113$ & $5.3420 \pm 0.0206$ \\
TabM     & $5.2084 \pm 0.0145$ & $5.1835 \pm 0.0121$ \\
TabICL   & $5.0034 \pm 0.0125$ & $4.9983 \pm 0.0071$ \\
\midrule
AutoPyTorch & \multicolumn{2}{c}{$5.5970$} \\
AutoKeras   & \multicolumn{2}{c}{$5.9986$} \\
\bottomrule
\end{tabular}
\end{minipage}

\end{center}

\Needspace{14\baselineskip}
\begin{center}
\scriptsize
\setlength{\tabcolsep}{4pt}
\renewcommand{\arraystretch}{1.04}

\begin{minipage}[t]{0.49\linewidth}
\centering
\textbf{house ($\times 10^{4}$) $\downarrow$}\\
\begin{tabular}{lcc}
\toprule
Model & Default HPO & Agentic HPO \\
\midrule
MLP      & $3.0997 \pm 0.0385$ & $3.0567 \pm 0.0429$ \\
RealMLP  & $3.0864 \pm 0.0286$ & $3.0734 \pm 0.0302$ \\
LightGBM & $3.1362 \pm 0.0031$ & $3.1339 \pm 0.0057$ \\
TabM     & $3.0441 \pm 0.0110$ & $2.9945 \pm 0.0101$ \\
TabICL   & $2.7962 \pm 0.0064$ & $2.8235 \pm 0.0000$ \\
\midrule
AutoPyTorch & \multicolumn{2}{c}{$3.1727$} \\
AutoKeras   & \multicolumn{2}{c}{$3.3025$} \\
\bottomrule
\end{tabular}
\end{minipage}
\hfill
\begin{minipage}[t]{0.49\linewidth}
\centering
\textbf{microsoft ($\times 10^{-1}$) $\downarrow$}\\
\begin{tabular}{lcc}
\toprule
Model & Default HPO & Agentic HPO \\
\midrule
MLP      & $7.4654 \pm 0.0036$ & $7.4610 \pm 0.0040$ \\
RealMLP  & $7.4176 \pm 0.0030$ & $7.4350 \pm 0.0017$ \\
LightGBM & $7.4159 \pm 0.0011$ & $7.4162 \pm 0.0011$ \\
TabM     & $7.4143 \pm 0.0020$ & $7.4285 \pm 0.0020$ \\
TabICL   & $7.6311 \pm 0.0248$ & $7.5985 \pm 0.0129$ \\
\midrule
AutoPyTorch & \multicolumn{2}{c}{$7.7543$} \\
AutoKeras   & \multicolumn{2}{c}{$7.6362$} \\
\bottomrule
\end{tabular}
\end{minipage}

\end{center}

\Needspace{14\baselineskip}
\begin{center}
\scriptsize
\setlength{\tabcolsep}{4pt}
\renewcommand{\arraystretch}{1.04}

\begin{minipage}[t]{0.49\linewidth}
\centering
\textbf{tabarena/airfoil\_self\_noise $\downarrow$}\\
\begin{tabular}{lcc}
\toprule
Model & Default HPO & Agentic HPO \\
\midrule
MLP      & $1.2111 \pm 0.0851$ & $1.0652 \pm 0.0188$ \\
RealMLP  & $1.1872 \pm 0.0386$ & $1.1341 \pm 0.0186$ \\
LightGBM & $1.6071 \pm 0.0331$ & $1.3727 \pm 0.0110$ \\
TabM     & $1.0746 \pm 0.0327$ & $1.1152 \pm 0.0335$ \\
TabICL   & $0.9991 \pm 0.0103$ & $1.0006 \pm 0.0119$ \\
\midrule
AutoPyTorch & \multicolumn{2}{c}{$1.9506$} \\
AutoKeras   & \multicolumn{2}{c}{$2.1980$} \\
\bottomrule
\end{tabular}
\end{minipage}
\hfill
\begin{minipage}[t]{0.49\linewidth}
\centering
\textbf{tabarena/Amazon\_employee\_access ($\times 10^{-1}$) $\uparrow$}\\
\begin{tabular}{lcc}
\toprule
Model & Default HPO & Agentic HPO \\
\midrule
MLP      & $8.3297 \pm 0.0155$ & $8.3507 \pm 0.0356$ \\
RealMLP  & $8.4477 \pm 0.0540$ & $8.5627 \pm 0.0634$ \\
LightGBM & $8.2752 \pm 0.0388$ & $8.3810 \pm 0.0258$ \\
TabM     & $8.3070 \pm 0.0772$ & $8.3013 \pm 0.0486$ \\
TabICL   & $8.4644 \pm 0.0119$ & $8.4433 \pm 0.0116$ \\
\midrule
AutoPyTorch & \multicolumn{2}{c}{$8.3501$} \\
AutoKeras   & \multicolumn{2}{c}{$6.0564$} \\
\bottomrule
\end{tabular}
\end{minipage}

\end{center}

\Needspace{14\baselineskip}
\begin{center}
\scriptsize
\setlength{\tabcolsep}{4pt}
\renewcommand{\arraystretch}{1.04}

\begin{minipage}[t]{0.49\linewidth}
\centering
\textbf{tabarena/Another-Dataset-on-used-Fiat-500 ($\times 10^{2}$) $\downarrow$}\\
\begin{tabular}{lcc}
\toprule
Model & Default HPO & Agentic HPO \\
\midrule
MLP      & $7.3823 \pm 0.0957$ & $7.1248 \pm 0.0164$ \\
RealMLP  & $7.4757 \pm 0.1191$ & $7.5793 \pm 0.1223$ \\
LightGBM & $7.1827 \pm 0.0270$ & $7.1960 \pm 0.0745$ \\
TabM     & $7.2767 \pm 0.0315$ & $7.1903 \pm 0.0398$ \\
TabICL   & $6.8200 \pm 0.0197$ & $6.8626 \pm 0.0056$ \\
\midrule
AutoPyTorch & \multicolumn{2}{c}{$7.1934$} \\
AutoKeras   & \multicolumn{2}{c}{$7.3005$} \\
\bottomrule
\end{tabular}
\end{minipage}
\hfill
\begin{minipage}[t]{0.49\linewidth}
\centering
\textbf{tabarena/APSFailure ($\times 10^{-1}$) $\uparrow$}\\
\begin{tabular}{lcc}
\toprule
Model & Default HPO & Agentic HPO \\
\midrule
MLP      & $9.9218 \pm 0.0064$ & $9.8721 \pm 0.0307$ \\
RealMLP  & $9.9309 \pm 0.0096$ & $9.9140 \pm 0.0140$ \\
LightGBM & $9.9287 \pm 0.0092$ & $9.9302 \pm 0.0080$ \\
TabM     & $9.9330 \pm 0.0062$ & $9.9426 \pm 0.0031$ \\
TabICL   & $9.9370 \pm 0.0017$ & $9.9389 \pm 0.0042$ \\
\midrule
AutoPyTorch & \multicolumn{2}{c}{$9.9031$} \\
AutoKeras   & \multicolumn{2}{c}{$9.8351$} \\
\bottomrule
\end{tabular}
\end{minipage}

\end{center}

\Needspace{14\baselineskip}
\begin{center}
\scriptsize
\setlength{\tabcolsep}{4pt}
\renewcommand{\arraystretch}{1.04}

\begin{minipage}[t]{0.49\linewidth}
\centering
\textbf{tabarena/bank-marketing ($\times 10^{-1}$) $\uparrow$}\\
\begin{tabular}{lcc}
\toprule
Model & Default HPO & Agentic HPO \\
\midrule
MLP      & $7.7260 \pm 0.0052$ & $7.6111 \pm 0.0182$ \\
RealMLP  & $7.6694 \pm 0.0149$ & $7.6746 \pm 0.0199$ \\
LightGBM & $7.7145 \pm 0.0166$ & $7.6781 \pm 0.0194$ \\
TabM     & $7.7285 \pm 0.0056$ & $7.6818 \pm 0.0093$ \\
TabICL   & $7.1941 \pm 0.3702$ & $7.4738 \pm 0.1351$ \\
\midrule
AutoPyTorch & \multicolumn{2}{c}{$7.4950$} \\
AutoKeras   & \multicolumn{2}{c}{$7.5013$} \\
\bottomrule
\end{tabular}
\end{minipage}
\hfill
\begin{minipage}[t]{0.49\linewidth}
\centering
\textbf{tabarena/Bioresponse ($\times 10^{-1}$) $\uparrow$}\\
\begin{tabular}{lcc}
\toprule
Model & Default HPO & Agentic HPO \\
\midrule
MLP      & $8.5302 \pm 0.0492$ & $8.5757 \pm 0.0368$ \\
RealMLP  & $8.6958 \pm 0.0415$ & $8.6243 \pm 0.0370$ \\
LightGBM & $8.7933 \pm 0.0377$ & $8.6900 \pm 0.0279$ \\
TabM     & $8.7259 \pm 0.0186$ & $8.6201 \pm 0.0194$ \\
TabICL   & $8.6475 \pm 0.0267$ & $8.6487 \pm 0.0116$ \\
\midrule
AutoPyTorch & \multicolumn{2}{c}{$8.2180$} \\
AutoKeras   & \multicolumn{2}{c}{$8.2771$} \\
\bottomrule
\end{tabular}
\end{minipage}

\end{center}

\Needspace{14\baselineskip}
\begin{center}
\scriptsize
\setlength{\tabcolsep}{4pt}
\renewcommand{\arraystretch}{1.04}

\begin{minipage}[t]{0.49\linewidth}
\centering
\textbf{tabarena/concrete\_compressive\_strength $\downarrow$}\\
\begin{tabular}{lcc}
\toprule
Model & Default HPO & Agentic HPO \\
\midrule
MLP      & $4.8574 \pm 0.1416$ & $4.6435 \pm 0.1284$ \\
RealMLP  & $4.6979 \pm 0.0916$ & $4.6595 \pm 0.0599$ \\
LightGBM & $4.7099 \pm 0.0469$ & $4.6605 \pm 0.1111$ \\
TabM     & $5.0098 \pm 0.1014$ & $4.3319 \pm 0.0715$ \\
TabICL   & $3.6744 \pm 0.0270$ & $3.6683 \pm 0.0307$ \\
\midrule
AutoPyTorch & \multicolumn{2}{c}{$5.3965$} \\
AutoKeras   & \multicolumn{2}{c}{$4.9451$} \\
\bottomrule
\end{tabular}
\end{minipage}
\hfill
\begin{minipage}[t]{0.49\linewidth}
\centering
\textbf{tabarena/credit\_card\_clients\_default ($\times 10^{-1}$) $\uparrow$}\\
\begin{tabular}{lcc}
\toprule
Model & Default HPO & Agentic HPO \\
\midrule
MLP      & $7.8489 \pm 0.0278$ & $7.8524 \pm 0.0232$ \\
RealMLP  & $7.9047 \pm 0.0109$ & $7.8049 \pm 0.0526$ \\
LightGBM & $7.8778 \pm 0.0116$ & $7.8909 \pm 0.0187$ \\
TabM     & $7.9234 \pm 0.0045$ & $7.8584 \pm 0.0132$ \\
TabICL   & $7.9457 \pm 0.0025$ & $7.9072 \pm 0.0137$ \\
\midrule
AutoPyTorch & \multicolumn{2}{c}{$7.8215$} \\
AutoKeras   & \multicolumn{2}{c}{$7.6977$} \\
\bottomrule
\end{tabular}
\end{minipage}

\end{center}

\Needspace{14\baselineskip}
\begin{center}
\scriptsize
\setlength{\tabcolsep}{4pt}
\renewcommand{\arraystretch}{1.04}

\begin{minipage}[t]{0.49\linewidth}
\centering
\textbf{tabarena/customer\_satisfaction\_in\_airline ($\times 10^{-1}$) $\uparrow$}\\
\begin{tabular}{lcc}
\toprule
Model & Default HPO & Agentic HPO \\
\midrule
MLP      & $9.9408 \pm 0.0007$ & $9.9488 \pm 0.0008$ \\
RealMLP  & $9.9511 \pm 0.0004$ & $9.9518 \pm 0.0007$ \\
LightGBM & $9.9415 \pm 0.0004$ & $9.9437 \pm 0.0003$ \\
TabM     & $9.9518 \pm 0.0006$ & $9.9535 \pm 0.0005$ \\
TabICL   & $9.9475 \pm 0.0003$ & $9.9476 \pm 0.0004$ \\
\midrule
AutoPyTorch & \multicolumn{2}{c}{$9.9261$} \\
AutoKeras   & \multicolumn{2}{c}{$9.8125$} \\
\bottomrule
\end{tabular}
\end{minipage}
\hfill
\begin{minipage}[t]{0.49\linewidth}
\centering
\textbf{tabarena/diabetes ($\times 10^{-1}$) $\uparrow$}\\
\begin{tabular}{lcc}
\toprule
Model & Default HPO & Agentic HPO \\
\midrule
MLP      & $8.1096 \pm 0.0801$ & $8.2129 \pm 0.1661$ \\
RealMLP  & $8.0019 \pm 0.2015$ & $8.2506 \pm 0.1144$ \\
LightGBM & $8.0315 \pm 0.1222$ & $7.9920 \pm 0.1538$ \\
TabM     & $7.9481 \pm 0.0844$ & $8.3092 \pm 0.0408$ \\
TabICL   & $8.4487 \pm 0.0036$ & $8.4329 \pm 0.0102$ \\
\midrule
AutoPyTorch & \multicolumn{2}{c}{$8.0877$} \\
AutoKeras   & \multicolumn{2}{c}{$8.1533$} \\
\bottomrule
\end{tabular}
\end{minipage}

\end{center}

\Needspace{14\baselineskip}
\begin{center}
\scriptsize
\setlength{\tabcolsep}{4pt}
\renewcommand{\arraystretch}{1.04}

\begin{minipage}[t]{0.49\linewidth}
\centering
\textbf{tabarena/Diabetes130US ($\times 10^{-1}$) $\uparrow$}\\
\begin{tabular}{lcc}
\toprule
Model & Default HPO & Agentic HPO \\
\midrule
MLP      & $6.6634 \pm 0.0100$ & $6.6510 \pm 0.0317$ \\
RealMLP  & $6.7708 \pm 0.0312$ & $6.5302 \pm 0.3399$ \\
LightGBM & $6.6224 \pm 0.0240$ & $6.6612 \pm 0.0247$ \\
TabM     & $6.6973 \pm 0.0296$ & $6.7207 \pm 0.0234$ \\
TabICL   & $6.6433 \pm 0.0999$ & $6.4517 \pm 0.0410$ \\
\midrule
AutoPyTorch & \multicolumn{2}{c}{$6.3983$} \\
AutoKeras   & \multicolumn{2}{c}{$6.0083$} \\
\bottomrule
\end{tabular}
\end{minipage}
\hfill
\begin{minipage}[t]{0.49\linewidth}
\centering
\textbf{tabarena/E-CommereShippingData ($\times 10^{-1}$) $\uparrow$}\\
\begin{tabular}{lcc}
\toprule
Model & Default HPO & Agentic HPO \\
\midrule
MLP      & $7.4061 \pm 0.0156$ & $7.3510 \pm 0.0548$ \\
RealMLP  & $7.4130 \pm 0.0404$ & $7.4087 \pm 0.0355$ \\
LightGBM & $7.4183 \pm 0.0089$ & $7.3755 \pm 0.0321$ \\
TabM     & $7.4042 \pm 0.0242$ & $7.3975 \pm 0.0193$ \\
TabICL   & $7.4065 \pm 0.0113$ & $7.3961 \pm 0.0144$ \\
\midrule
AutoPyTorch & \multicolumn{2}{c}{$7.3723$} \\
AutoKeras   & \multicolumn{2}{c}{$7.3439$} \\
\bottomrule
\end{tabular}
\end{minipage}

\end{center}

\Needspace{14\baselineskip}
\begin{center}
\scriptsize
\setlength{\tabcolsep}{4pt}
\renewcommand{\arraystretch}{1.04}

\begin{minipage}[t]{0.49\linewidth}
\centering
\textbf{tabarena/Food\_Delivery\_Time $\downarrow$}\\
\begin{tabular}{lcc}
\toprule
Model & Default HPO & Agentic HPO \\
\midrule
MLP      & $7.6184 \pm 0.0065$ & $7.3969 \pm 0.0129$ \\
RealMLP  & $7.4429 \pm 0.0144$ & $7.4876 \pm 0.0108$ \\
LightGBM & $7.5184 \pm 0.0024$ & $7.2845 \pm 0.0011$ \\
TabM     & $7.7018 \pm 0.0174$ & $7.3772 \pm 0.0059$ \\
TabICL   & $7.6786 \pm 0.0226$ & $7.4968 \pm 0.0000$ \\
\midrule
AutoPyTorch & \multicolumn{2}{c}{$7.7677$} \\
AutoKeras   & \multicolumn{2}{c}{$8.0159$} \\
\bottomrule
\end{tabular}
\end{minipage}
\hfill
\begin{minipage}[t]{0.49\linewidth}
\centering
\textbf{tabarena/GiveMeSomeCredit ($\times 10^{-1}$) $\uparrow$}\\
\begin{tabular}{lcc}
\toprule
Model & Default HPO & Agentic HPO \\
\midrule
MLP      & $8.7016 \pm 0.0023$ & $8.7004 \pm 0.0023$ \\
RealMLP  & $8.7012 \pm 0.0038$ & $8.6886 \pm 0.0042$ \\
LightGBM & $8.6948 \pm 0.0012$ & $8.6812 \pm 0.0014$ \\
TabM     & $8.6901 \pm 0.0020$ & $8.6940 \pm 0.0035$ \\
TabICL   & $8.6447 \pm 0.1262$ & $8.6993 \pm 0.0023$ \\
\midrule
AutoPyTorch & \multicolumn{2}{c}{$8.5903$} \\
AutoKeras   & \multicolumn{2}{c}{$8.3068$} \\
\bottomrule
\end{tabular}
\end{minipage}

\end{center}

\Needspace{14\baselineskip}
\begin{center}
\scriptsize
\setlength{\tabcolsep}{4pt}
\renewcommand{\arraystretch}{1.04}

\begin{minipage}[t]{0.49\linewidth}
\centering
\textbf{tabarena/healthcare\_insurance\_expenses ($\times 10^{3}$) $\downarrow$}\\
\begin{tabular}{lcc}
\toprule
Model & Default HPO & Agentic HPO \\
\midrule
MLP      & $4.3154 \pm 0.0611$ & $4.3432 \pm 0.0585$ \\
RealMLP  & $4.2618 \pm 0.0441$ & $4.1889 \pm 0.0545$ \\
LightGBM & $4.2711 \pm 0.0252$ & $4.3164 \pm 0.1070$ \\
TabM     & $4.1117 \pm 0.0085$ & $4.1942 \pm 0.0428$ \\
TabICL   & $4.0512 \pm 0.0046$ & $4.0887 \pm 0.0328$ \\
\midrule
AutoPyTorch & \multicolumn{2}{c}{$4.6827$} \\
AutoKeras   & \multicolumn{2}{c}{$4.3969$} \\
\bottomrule
\end{tabular}
\end{minipage}
\hfill
\begin{minipage}[t]{0.49\linewidth}
\centering
\textbf{tabarena/heloc ($\times 10^{-1}$) $\uparrow$}\\
\begin{tabular}{lcc}
\toprule
Model & Default HPO & Agentic HPO \\
\midrule
MLP      & $7.9526 \pm 0.0074$ & $7.9523 \pm 0.0056$ \\
RealMLP  & $7.9769 \pm 0.0021$ & $7.9419 \pm 0.0180$ \\
LightGBM & $7.9061 \pm 0.0186$ & $7.9584 \pm 0.0066$ \\
TabM     & $7.9618 \pm 0.0059$ & $7.9752 \pm 0.0056$ \\
TabICL   & $8.0047 \pm 0.0072$ & $8.0102 \pm 0.0042$ \\
\midrule
AutoPyTorch & \multicolumn{2}{c}{$7.8953$} \\
AutoKeras   & \multicolumn{2}{c}{$7.8187$} \\
\bottomrule
\end{tabular}
\end{minipage}

\end{center}

\Needspace{14\baselineskip}
\begin{center}
\scriptsize
\setlength{\tabcolsep}{4pt}
\renewcommand{\arraystretch}{1.04}

\begin{minipage}[t]{0.49\linewidth}
\centering
\textbf{tabarena/HR\_Analytics\_Job\_Change\_of\_Data\_Scientists ($\times 10^{-1}$) $\uparrow$}\\
\begin{tabular}{lcc}
\toprule
Model & Default HPO & Agentic HPO \\
\midrule
MLP      & $8.0911 \pm 0.0111$ & $8.0883 \pm 0.0096$ \\
RealMLP  & $8.1093 \pm 0.0160$ & $8.1034 \pm 0.0218$ \\
LightGBM & $8.0685 \pm 0.0134$ & $8.0531 \pm 0.0221$ \\
TabM     & $8.1173 \pm 0.0068$ & $8.1175 \pm 0.0060$ \\
TabICL   & $8.0661 \pm 0.0627$ & $8.1450 \pm 0.0100$ \\
\midrule
AutoPyTorch & \multicolumn{2}{c}{$8.0030$} \\
AutoKeras   & \multicolumn{2}{c}{$7.9288$} \\
\bottomrule
\end{tabular}
\end{minipage}
\hfill
\begin{minipage}[t]{0.49\linewidth}
\centering
\textbf{tabarena/in\_vehicle\_coupon\_recommendation ($\times 10^{-1}$) $\uparrow$}\\
\begin{tabular}{lcc}
\toprule
Model & Default HPO & Agentic HPO \\
\midrule
MLP      & $8.2448 \pm 0.0308$ & $8.2877 \pm 0.0258$ \\
RealMLP  & $8.4424 \pm 0.0155$ & $8.4188 \pm 0.0199$ \\
LightGBM & $8.4402 \pm 0.0067$ & $8.4756 \pm 0.0056$ \\
TabM     & $8.5234 \pm 0.0183$ & $8.5315 \pm 0.0154$ \\
TabICL   & $8.4973 \pm 0.0291$ & $8.4963 \pm 0.0020$ \\
\midrule
AutoPyTorch & \multicolumn{2}{c}{$8.2578$} \\
AutoKeras   & \multicolumn{2}{c}{$7.3155$} \\
\bottomrule
\end{tabular}
\end{minipage}

\end{center}

\Needspace{14\baselineskip}
\begin{center}
\scriptsize
\setlength{\tabcolsep}{4pt}
\renewcommand{\arraystretch}{1.04}

\begin{minipage}[t]{0.49\linewidth}
\centering
\textbf{tabarena/jm1 ($\times 10^{-1}$) $\uparrow$}\\
\begin{tabular}{lcc}
\toprule
Model & Default HPO & Agentic HPO \\
\midrule
MLP      & $7.3112 \pm 0.0433$ & $7.3877 \pm 0.0272$ \\
RealMLP  & $7.4373 \pm 0.0399$ & $7.4161 \pm 0.0271$ \\
LightGBM & $7.5327 \pm 0.0232$ & $7.4334 \pm 0.0633$ \\
TabM     & $7.3276 \pm 0.0180$ & $7.3322 \pm 0.0387$ \\
TabICL   & $7.4488 \pm 0.1107$ & $7.7270 \pm 0.0740$ \\
\midrule
AutoPyTorch & \multicolumn{2}{c}{$7.3836$} \\
AutoKeras   & \multicolumn{2}{c}{$7.3327$} \\
\bottomrule
\end{tabular}
\end{minipage}
\hfill
\begin{minipage}[t]{0.49\linewidth}
\centering
\textbf{tabarena/kddcup09\_appetency ($\times 10^{-1}$) $\uparrow$}\\
\begin{tabular}{lcc}
\toprule
Model & Default HPO & Agentic HPO \\
\midrule
MLP      & $8.1465 \pm 0.0365$ & $8.0903 \pm 0.0258$ \\
RealMLP  & $8.1760 \pm 0.0284$ & $7.6745 \pm 0.5382$ \\
LightGBM & $8.0864 \pm 0.0303$ & $8.1256 \pm 0.0408$ \\
TabM     & $8.1329 \pm 0.0259$ & $8.1542 \pm 0.0302$ \\
TabICL   & $8.1179 \pm 0.0126$ & $8.1086 \pm 0.0134$ \\
\midrule
AutoPyTorch & \multicolumn{2}{c}{$6.2612$} \\
AutoKeras   & \multicolumn{2}{c}{$7.5823$} \\
\bottomrule
\end{tabular}
\end{minipage}

\end{center}

\Needspace{14\baselineskip}
\begin{center}
\scriptsize
\setlength{\tabcolsep}{4pt}
\renewcommand{\arraystretch}{1.04}

\begin{minipage}[t]{0.49\linewidth}
\centering
\textbf{tabarena/miami\_housing ($\times 10^{4}$) $\downarrow$}\\
\begin{tabular}{lcc}
\toprule
Model & Default HPO & Agentic HPO \\
\midrule
MLP      & $8.8229 \pm 0.0985$ & $8.6732 \pm 0.1253$ \\
RealMLP  & $8.4009 \pm 0.1724$ & $8.3701 \pm 0.0671$ \\
LightGBM & $9.1026 \pm 0.0869$ & $9.0188 \pm 0.0621$ \\
TabM     & $8.2868 \pm 0.0378$ & $8.3488 \pm 0.1302$ \\
TabICL   & $7.9680 \pm 0.0340$ & $7.9111 \pm 0.0163$ \\
\midrule
AutoPyTorch & \multicolumn{2}{c}{$9.2342$} \\
AutoKeras   & \multicolumn{2}{c}{$9.0830$} \\
\bottomrule
\end{tabular}
\end{minipage}
\hfill
\begin{minipage}[t]{0.49\linewidth}
\centering
\textbf{tabarena/NATICUSdroid ($\times 10^{-1}$) $\uparrow$}\\
\begin{tabular}{lcc}
\toprule
Model & Default HPO & Agentic HPO \\
\midrule
MLP      & $9.8505 \pm 0.0033$ & $9.8331 \pm 0.0107$ \\
RealMLP  & $9.8443 \pm 0.0050$ & $9.8460 \pm 0.0055$ \\
LightGBM & $9.8538 \pm 0.0013$ & $9.8442 \pm 0.0026$ \\
TabM     & $9.8509 \pm 0.0019$ & $9.8461 \pm 0.0055$ \\
TabICL   & $9.8665 \pm 0.0099$ & $9.8580 \pm 0.0148$ \\
\midrule
AutoPyTorch & \multicolumn{2}{c}{$9.8165$} \\
AutoKeras   & \multicolumn{2}{c}{$9.8083$} \\
\bottomrule
\end{tabular}
\end{minipage}

\end{center}

\Needspace{14\baselineskip}
\begin{center}
\scriptsize
\setlength{\tabcolsep}{4pt}
\renewcommand{\arraystretch}{1.04}

\begin{minipage}[t]{0.49\linewidth}
\centering
\textbf{tabarena/online\_shoppers\_intention ($\times 10^{-1}$) $\uparrow$}\\
\begin{tabular}{lcc}
\toprule
Model & Default HPO & Agentic HPO \\
\midrule
MLP      & $9.2454 \pm 0.0153$ & $9.2592 \pm 0.0099$ \\
RealMLP  & $9.2977 \pm 0.0135$ & $9.2838 \pm 0.0078$ \\
LightGBM & $9.2484 \pm 0.0235$ & $9.2670 \pm 0.0036$ \\
TabM     & $9.2785 \pm 0.0121$ & $9.2689 \pm 0.0051$ \\
TabICL   & $9.3245 \pm 0.0117$ & $9.3004 \pm 0.0157$ \\
\midrule
AutoPyTorch & \multicolumn{2}{c}{$9.2502$} \\
AutoKeras   & \multicolumn{2}{c}{$8.9242$} \\
\bottomrule
\end{tabular}
\end{minipage}
\hfill
\begin{minipage}[t]{0.49\linewidth}
\centering
\textbf{tabarena/physiochemical\_protein $\downarrow$}\\
\begin{tabular}{lcc}
\toprule
Model & Default HPO & Agentic HPO \\
\midrule
MLP      & $3.5443 \pm 0.0155$ & $3.6026 \pm 0.0139$ \\
RealMLP  & $3.2988 \pm 0.0134$ & $3.2711 \pm 0.0112$ \\
LightGBM & $3.5027 \pm 0.0022$ & $3.2991 \pm 0.0044$ \\
TabM     & $3.3609 \pm 0.0255$ & $3.3444 \pm 0.0138$ \\
TabICL   & $3.0133 \pm 0.0066$ & $3.0075 \pm 0.0040$ \\
\midrule
AutoPyTorch & \multicolumn{2}{c}{$3.6083$} \\
AutoKeras   & \multicolumn{2}{c}{$3.8070$} \\
\bottomrule
\end{tabular}
\end{minipage}

\end{center}

\Needspace{14\baselineskip}
\begin{center}
\scriptsize
\setlength{\tabcolsep}{4pt}
\renewcommand{\arraystretch}{1.04}

\begin{minipage}[t]{0.49\linewidth}
\centering
\textbf{tabarena/polish\_companies\_bankruptcy ($\times 10^{-1}$) $\uparrow$}\\
\begin{tabular}{lcc}
\toprule
Model & Default HPO & Agentic HPO \\
\midrule
MLP      & $9.2275 \pm 0.0604$ & $9.3093 \pm 0.0684$ \\
RealMLP  & $9.3074 \pm 0.0902$ & $9.5427 \pm 0.0418$ \\
LightGBM & $9.5801 \pm 0.0271$ & $9.6194 \pm 0.0178$ \\
TabM     & $9.5876 \pm 0.0238$ & $9.2369 \pm 0.0555$ \\
TabICL   & $9.8386 \pm 0.0074$ & $9.8576 \pm 0.0073$ \\
\midrule
AutoPyTorch & \multicolumn{2}{c}{$9.3865$} \\
AutoKeras   & \multicolumn{2}{c}{$8.6602$} \\
\bottomrule
\end{tabular}
\end{minipage}
\hfill
\begin{minipage}[t]{0.49\linewidth}
\centering
\textbf{tabarena/qsar-biodeg ($\times 10^{-1}$) $\uparrow$}\\
\begin{tabular}{lcc}
\toprule
Model & Default HPO & Agentic HPO \\
\midrule
MLP      & $9.1409 \pm 0.0369$ & $9.1758 \pm 0.1390$ \\
RealMLP  & $9.1275 \pm 0.0500$ & $9.2730 \pm 0.0261$ \\
LightGBM & $9.0573 \pm 0.0837$ & $9.0792 \pm 0.0891$ \\
TabM     & $9.3032 \pm 0.0186$ & $9.2689 \pm 0.0367$ \\
TabICL   & $9.3723 \pm 0.0072$ & $9.3363 \pm 0.0088$ \\
\midrule
AutoPyTorch & \multicolumn{2}{c}{$8.9202$} \\
AutoKeras   & \multicolumn{2}{c}{$9.2520$} \\
\bottomrule
\end{tabular}
\end{minipage}

\end{center}

\Needspace{14\baselineskip}
\begin{center}
\scriptsize
\setlength{\tabcolsep}{4pt}
\renewcommand{\arraystretch}{1.04}

\begin{minipage}[t]{0.49\linewidth}
\centering
\textbf{tabarena/QSAR-TID-11 ($\times 10^{-1}$) $\downarrow$}\\
\begin{tabular}{lcc}
\toprule
Model & Default HPO & Agentic HPO \\
\midrule
MLP      & $8.8851 \pm 0.0623$ & $8.3229 \pm 0.0459$ \\
RealMLP  & $8.4800 \pm 0.0718$ & $8.5389 \pm 0.0369$ \\
LightGBM & $8.3262 \pm 0.0274$ & $8.2736 \pm 0.0224$ \\
TabM     & $8.4739 \pm 0.0186$ & $8.2702 \pm 0.0289$ \\
TabICL   & $8.7313 \pm 0.0534$ & $8.7626 \pm 0.0475$ \\
\midrule
AutoPyTorch & \multicolumn{2}{c}{$8.8098$} \\
AutoKeras   & \multicolumn{2}{c}{$9.1580$} \\
\bottomrule
\end{tabular}
\end{minipage}
\hfill
\begin{minipage}[t]{0.49\linewidth}
\centering
\textbf{tabarena/QSAR\_fish\_toxicity ($\times 10^{-1}$) $\downarrow$}\\
\begin{tabular}{lcc}
\toprule
Model & Default HPO & Agentic HPO \\
\midrule
MLP      & $9.8934 \pm 0.3439$ & $9.7175 \pm 0.1754$ \\
RealMLP  & $9.2329 \pm 0.0780$ & $9.4535 \pm 0.1564$ \\
LightGBM & $9.6557 \pm 0.1513$ & $9.6247 \pm 0.1816$ \\
TabM     & $9.3465 \pm 0.0941$ & $9.4143 \pm 0.0778$ \\
TabICL   & $8.9515 \pm 0.0132$ & $8.9564 \pm 0.0000$ \\
\midrule
AutoPyTorch & \multicolumn{2}{c}{$9.5082$} \\
AutoKeras   & \multicolumn{2}{c}{$9.8908$} \\
\bottomrule
\end{tabular}
\end{minipage}

\end{center}

\Needspace{14\baselineskip}
\begin{center}
\scriptsize
\setlength{\tabcolsep}{4pt}
\renewcommand{\arraystretch}{1.04}

\begin{minipage}[t]{0.49\linewidth}
\centering
\textbf{tabarena/splice ($\times 10^{-1}$) $\downarrow$}\\
\begin{tabular}{lcc}
\toprule
Model & Default HPO & Agentic HPO \\
\midrule
MLP      & $1.1486 \pm 0.0190$ & $0.9329 \pm 0.0475$ \\
RealMLP  & $1.0274 \pm 0.0459$ & $1.0824 \pm 0.5004$ \\
LightGBM & $1.0272 \pm 0.0422$ & $0.9034 \pm 0.0424$ \\
TabM     & $0.9576 \pm 0.0111$ & $1.0152 \pm 0.0406$ \\
TabICL   & $0.7346 \pm 0.0099$ & $0.7314 \pm 0.0058$ \\
\midrule
AutoPyTorch & \multicolumn{2}{c}{$1.4024$} \\
AutoKeras   & \multicolumn{2}{c}{$1.9949$} \\
\bottomrule
\end{tabular}
\end{minipage}
\hfill
\begin{minipage}[t]{0.49\linewidth}
\centering
\textbf{tabarena/taiwanese\_bankruptcy\_prediction ($\times 10^{-1}$) $\uparrow$}\\
\begin{tabular}{lcc}
\toprule
Model & Default HPO & Agentic HPO \\
\midrule
MLP      & $9.3664 \pm 0.0071$ & $8.9853 \pm 0.1144$ \\
RealMLP  & $9.2386 \pm 0.1752$ & $9.2692 \pm 0.2576$ \\
LightGBM & $9.4283 \pm 0.0450$ & $9.3512 \pm 0.1165$ \\
TabM     & $9.4117 \pm 0.0177$ & $9.4787 \pm 0.0163$ \\
TabICL   & $9.4728 \pm 0.0077$ & $9.5173 \pm 0.0067$ \\
\midrule
AutoPyTorch & \multicolumn{2}{c}{$9.1591$} \\
AutoKeras   & \multicolumn{2}{c}{$8.7867$} \\
\bottomrule
\end{tabular}
\end{minipage}

\end{center}

\Needspace{14\baselineskip}
\begin{center}
\scriptsize
\setlength{\tabcolsep}{4pt}
\renewcommand{\arraystretch}{1.04}

\begin{minipage}[t]{0.49\linewidth}
\centering
\textbf{tabarena/wine\_quality ($\times 10^{-1}$) $\downarrow$}\\
\begin{tabular}{lcc}
\toprule
Model & Default HPO & Agentic HPO \\
\midrule
MLP      & $6.7464 \pm 0.0426$ & $6.5243 \pm 0.0594$ \\
RealMLP  & $6.0529 \pm 0.0300$ & $6.0895 \pm 0.0278$ \\
LightGBM & $6.1508 \pm 0.0125$ & $6.0856 \pm 0.0196$ \\
TabM     & $6.3776 \pm 0.0724$ & $6.2001 \pm 0.0247$ \\
TabICL   & $5.8641 \pm 0.0064$ & $5.8630 \pm 0.0047$ \\
\midrule
AutoPyTorch & \multicolumn{2}{c}{$6.6550$} \\
AutoKeras   & \multicolumn{2}{c}{$6.6474$} \\
\bottomrule
\end{tabular}
\end{minipage}
\hfill
\begin{minipage}[t]{0.49\linewidth}
\centering
\textbf{tabred/cooking-time ($\times 10^{-1}$) $\downarrow$}\\
\begin{tabular}{lcc}
\toprule
Model & Default HPO & Agentic HPO \\
\midrule
MLP      & $4.8084 \pm 0.0041$ & $4.8119 \pm 0.0025$ \\
RealMLP  & $4.8131 \pm 0.0055$ & $4.8035 \pm 0.0029$ \\
LightGBM & $4.8248 \pm 0.0009$ & $4.8235 \pm 0.0008$ \\
TabM     & $4.8029 \pm 0.0037$ & $4.8022 \pm 0.0033$ \\
TabICL   & $4.8954 \pm 0.0056$ & $4.8828 \pm 0.0059$ \\
\midrule
AutoPyTorch & \multicolumn{2}{c}{$4.8263$} \\
AutoKeras   & \multicolumn{2}{c}{$4.9050$} \\
\bottomrule
\end{tabular}
\end{minipage}

\end{center}

\Needspace{14\baselineskip}
\begin{center}
\scriptsize
\setlength{\tabcolsep}{4pt}
\renewcommand{\arraystretch}{1.04}

\begin{minipage}[t]{0.49\linewidth}
\centering
\textbf{tabred/delivery-eta ($\times 10^{-1}$) $\downarrow$}\\
\begin{tabular}{lcc}
\toprule
Model & Default HPO & Agentic HPO \\
\midrule
MLP      & $5.5289 \pm 0.0164$ & $5.5336 \pm 0.0197$ \\
RealMLP  & $5.4760 \pm 0.0096$ & $5.4708 \pm 0.0102$ \\
LightGBM & $5.4694 \pm 0.0016$ & $5.4654 \pm 0.0010$ \\
TabM     & $5.5013 \pm 0.0167$ & $5.4949 \pm 0.0092$ \\
TabICL   & $5.5830 \pm 0.0061$ & $5.5812 \pm 0.0047$ \\
\midrule
AutoPyTorch & \multicolumn{2}{c}{$5.5704$} \\
AutoKeras   & \multicolumn{2}{c}{$5.5888$} \\
\bottomrule
\end{tabular}
\end{minipage}
\hfill
\begin{minipage}[t]{0.49\linewidth}
\centering
\textbf{tabred/ecom-offers ($\times 10^{-1}$) $\uparrow$}\\
\begin{tabular}{lcc}
\toprule
Model & Default HPO & Agentic HPO \\
\midrule
MLP      & $5.9667 \pm 0.0181$ & $6.0484 \pm 0.0138$ \\
RealMLP  & $5.9346 \pm 0.0202$ & $5.8116 \pm 0.1743$ \\
LightGBM & $5.7377 \pm 0.0326$ & $5.7700 \pm 0.0565$ \\
TabM     & $5.8804 \pm 0.0146$ & $6.0097 \pm 0.0137$ \\
TabICL   & $5.8372 \pm 0.1184$ & $5.9924 \pm 0.0298$ \\
\midrule
AutoPyTorch & \multicolumn{2}{c}{$6.1761$} \\
AutoKeras   & \multicolumn{2}{c}{$5.8387$} \\
\bottomrule
\end{tabular}
\end{minipage}

\end{center}

\Needspace{14\baselineskip}
\begin{center}
\scriptsize
\setlength{\tabcolsep}{4pt}
\renewcommand{\arraystretch}{1.04}

\begin{minipage}[t]{0.49\linewidth}
\centering
\textbf{tabred/homecredit-default ($\times 10^{-1}$) $\uparrow$}\\
\begin{tabular}{lcc}
\toprule
Model & Default HPO & Agentic HPO \\
\midrule
MLP      & $8.5875 \pm 0.0137$ & $8.5842 \pm 0.0103$ \\
RealMLP  & $8.6299 \pm 0.0130$ & $8.6260 \pm 0.0051$ \\
LightGBM & $8.6790 \pm 0.0033$ & $8.6757 \pm 0.0031$ \\
TabM     & $8.6438 \pm 0.0071$ & $8.6353 \pm 0.0046$ \\
TabICL   & $8.0993 \pm 0.0194$ & $8.0900 \pm 0.0161$ \\
\midrule
AutoPyTorch & \multicolumn{2}{c}{$6.3470$} \\
AutoKeras   & \multicolumn{2}{c}{$5.8340$} \\
\bottomrule
\end{tabular}
\end{minipage}
\hfill
\begin{minipage}[t]{0.49\linewidth}
\centering
\textbf{tabred/homesite-insurance ($\times 10^{-1}$) $\uparrow$}\\
\begin{tabular}{lcc}
\toprule
Model & Default HPO & Agentic HPO \\
\midrule
MLP      & $9.5994 \pm 0.0077$ & $9.6137 \pm 0.0052$ \\
RealMLP  & $9.6511 \pm 0.0032$ & $9.6329 \pm 0.0046$ \\
LightGBM & $9.6063 \pm 0.0017$ & $9.6111 \pm 0.0029$ \\
TabM     & $9.6246 \pm 0.0034$ & $9.6297 \pm 0.0017$ \\
TabICL   & $9.4678 \pm 0.0351$ & $9.5267 \pm 0.0076$ \\
\midrule
AutoPyTorch & \multicolumn{2}{c}{$9.4293$} \\
AutoKeras   & \multicolumn{2}{c}{$9.3546$} \\
\bottomrule
\end{tabular}
\end{minipage}

\end{center}

\Needspace{14\baselineskip}
\begin{center}
\scriptsize
\setlength{\tabcolsep}{4pt}
\renewcommand{\arraystretch}{1.04}

\begin{minipage}[t]{0.49\linewidth}
\centering
\textbf{tabred/maps-routing ($\times 10^{-1}$) $\downarrow$}\\
\begin{tabular}{lcc}
\toprule
Model & Default HPO & Agentic HPO \\
\midrule
MLP      & $1.6176 \pm 0.0018$ & $1.6207 \pm 0.0020$ \\
RealMLP  & $1.6117 \pm 0.0021$ & $1.6089 \pm 0.0015$ \\
LightGBM & $1.6165 \pm 0.0004$ & $1.6166 \pm 0.0004$ \\
TabM     & $1.6100 \pm 0.0012$ & $1.6101 \pm 0.0008$ \\
TabICL   & $1.6886 \pm 0.0026$ & $1.6897 \pm 0.0016$ \\
\midrule
AutoPyTorch & \multicolumn{2}{c}{$1.6537$} \\
AutoKeras   & \multicolumn{2}{c}{$1.7349$} \\
\bottomrule
\end{tabular}
\end{minipage}
\hfill
\begin{minipage}[t]{0.49\linewidth}
\centering
\textbf{tabred/sberbank-housing ($\times 10^{-1}$) $\downarrow$}\\
\begin{tabular}{lcc}
\toprule
Model & Default HPO & Agentic HPO \\
\midrule
MLP      & $2.4056 \pm 0.0251$ & $2.3904 \pm 0.0285$ \\
RealMLP  & $2.3101 \pm 0.0090$ & $2.3578 \pm 0.0232$ \\
LightGBM & $2.4882 \pm 0.0154$ & $2.5381 \pm 0.0208$ \\
TabM     & $2.3646 \pm 0.0235$ & $2.3863 \pm 0.0174$ \\
TabICL   & $2.8047 \pm 0.0498$ & $2.4961 \pm 0.0145$ \\
\midrule
AutoPyTorch & \multicolumn{2}{c}{$3.6158$} \\
AutoKeras   & \multicolumn{2}{c}{$5.4295$} \\
\bottomrule
\end{tabular}
\end{minipage}

\end{center}

\Needspace{14\baselineskip}
\begin{center}
\scriptsize
\setlength{\tabcolsep}{4pt}
\renewcommand{\arraystretch}{1.04}

\begin{minipage}[t]{0.49\linewidth}
\centering
\textbf{tabred/weather $\downarrow$}\\
\begin{tabular}{lcc}
\toprule
Model & Default HPO & Agentic HPO \\
\midrule
MLP      & $1.5212 \pm 0.0032$ & $1.5136 \pm 0.0021$ \\
RealMLP  & $1.4363 \pm 0.0020$ & $1.4389 \pm 0.0016$ \\
LightGBM & $1.4619 \pm 0.0009$ & $1.4657 \pm 0.0009$ \\
TabM     & $1.4669 \pm 0.0025$ & $1.4654 \pm 0.0021$ \\
TabICL   & $1.5829 \pm 0.0045$ & $1.5670 \pm 0.0024$ \\
\midrule
AutoPyTorch & \multicolumn{2}{c}{$1.5236$} \\
AutoKeras   & \multicolumn{2}{c}{$1.6028$} \\
\bottomrule
\end{tabular}
\end{minipage}

\end{center}

\subsection{Ensemble results}
\label{app:ensemble-per-dataset}

Table~\ref{tab:ensemble_per_dataset_results} reports the per-dataset greedy-ensemble results behind the ensembling discussion in Section~\ref{sec:main-performance}. Each subtable lists the four families (MLP$^\dagger$, TabM$^\dagger$, RealMLP, and LightGBM) with their base and agentic search spaces, and for each (family, space) reports three quantities: the ensemble's test metric (\emph{Ens.}), the test metric of the single strongest ensemble member (\emph{Best}), and the mean pairwise test-prediction correlation across the selected members ($\rho$; lower $\rightarrow$ more diverse pool). Ensembles are constructed by greedy forward selection following~\citet{caruana2004ensemble} on a pool of $100$ ($20$ for the large datasets) random-hyperparameter configurations per (family, space). In the few cases where greedy selection collapses to a single distinct member, the pairwise correlation is undefined, which we denote by ``--'' in the table. Below the family rows, each subtable lists the ensemble rows of the AutoPytorch and AutoGluon (see Appendix~\ref{app:automl-details} for details). We include the default AutoGluon portfolio under \emph{Default} column, and AutoGluon with agentic RealMLP and TabM$^\dagger$ under \emph{Agentic}. Arrows indicate whether higher or lower values are better.

\Needspace{18\baselineskip}
\begin{center}
\captionof{table}{Per-dataset greedy-ensemble results for each of the 45 datasets. Each subtable reports, for the default and agentic search spaces of every model family, the ensemble test metric (Ens.), the test metric of its strongest single member (Best), and the mean pairwise prediction correlation among selected members ($\rho$). Arrows indicate whether higher or lower values are better.}
\label{tab:ensemble_per_dataset_results}
\end{center}

\Needspace{14\baselineskip}
\begin{center}
\scriptsize
\setlength{\tabcolsep}{2pt}
\renewcommand{\arraystretch}{1.04}

\begin{minipage}[t]{0.49\linewidth}
\centering
\textbf{black-friday ($\times 10^{3}$) $\downarrow$}\\
\begin{tabular}{l@{\hskip 3pt}ccc@{\hskip 5pt}ccc}
\toprule
 & \multicolumn{3}{c}{Default} & \multicolumn{3}{c}{Agentic} \\
\cmidrule(lr){2-4}\cmidrule(lr){5-7}
Model & Ens. & Best & $\rho$ & Ens. & Best & $\rho$ \\
\midrule
MLP$^\dagger$ & 3.4609 & 3.4762 & 0.980 & 3.4595 & 3.4787 & 0.963 \\
TabM$^\dagger$ & 3.4565 & 3.4602 & 0.983 & 3.4599 & 3.4645 & 0.966 \\
LightGBM & 3.4538 & 3.4558 & 0.982 & 3.4540 & 3.4645 & 0.947 \\
RealMLP & 3.4458 & 3.4541 & 0.992 & 3.4432 & 3.4471 & 0.976 \\
\midrule
AutoPyTorch & 3.4638 & -- & -- & -- & -- & -- \\
AutoGluon & 3.4432 & -- & -- & 3.4417 & -- & -- \\
\bottomrule
\end{tabular}
\end{minipage}
\hfill
\begin{minipage}[t]{0.49\linewidth}
\centering
\textbf{california ($\times 10^{-1}$) $\downarrow$}\\
\begin{tabular}{l@{\hskip 3pt}ccc@{\hskip 5pt}ccc}
\toprule
 & \multicolumn{3}{c}{Default} & \multicolumn{3}{c}{Agentic} \\
\cmidrule(lr){2-4}\cmidrule(lr){5-7}
Model & Ens. & Best & $\rho$ & Ens. & Best & $\rho$ \\
\midrule
MLP$^\dagger$ & 4.4171 & 4.5278 & 0.980 & 4.3439 & 4.5809 & 0.947 \\
TabM$^\dagger$ & 4.1570 & 4.2247 & 0.985 & 4.0790 & 4.2081 & 0.980 \\
LightGBM & 4.2722 & 4.3053 & 0.967 & 4.3162 & 4.4208 & 0.978 \\
RealMLP & 4.0541 & 4.1375 & 0.983 & 3.9219 & 3.9471 & 0.982 \\
\midrule
AutoPyTorch & 4.2516 & -- & -- & -- & -- & -- \\
AutoGluon & 4.0306 & -- & -- & 3.9467 & -- & -- \\
\bottomrule
\end{tabular}
\end{minipage}

\end{center}

\Needspace{14\baselineskip}
\begin{center}
\scriptsize
\setlength{\tabcolsep}{2pt}
\renewcommand{\arraystretch}{1.04}

\begin{minipage}[t]{0.49\linewidth}
\centering
\textbf{churn ($\times 10^{-1}$) $\uparrow$}\\
\begin{tabular}{l@{\hskip 3pt}ccc@{\hskip 5pt}ccc}
\toprule
 & \multicolumn{3}{c}{Default} & \multicolumn{3}{c}{Agentic} \\
\cmidrule(lr){2-4}\cmidrule(lr){5-7}
Model & Ens. & Best & $\rho$ & Ens. & Best & $\rho$ \\
\midrule
MLP$^\dagger$ & 8.6100 & 8.5750 & 0.951 & 8.6300 & 8.6500 & 0.884 \\
TabM$^\dagger$ & 8.5750 & 8.5700 & 0.867 & 8.5800 & 8.5850 & 0.856 \\
LightGBM & 8.6100 & 8.5750 & 0.960 & 8.6250 & 8.5600 & 0.409 \\
RealMLP & 8.6200 & 8.5600 & 0.986 & 8.6100 & 8.6250 & 0.984 \\
\midrule
AutoPyTorch & 8.6000 & -- & -- & -- & -- & -- \\
AutoGluon & 8.6050 & -- & -- & 8.6550 & -- & -- \\
\bottomrule
\end{tabular}
\end{minipage}
\hfill
\begin{minipage}[t]{0.49\linewidth}
\centering
\textbf{diamond ($\times 10^{2}$) $\downarrow$}\\
\begin{tabular}{l@{\hskip 3pt}ccc@{\hskip 5pt}ccc}
\toprule
 & \multicolumn{3}{c}{Default} & \multicolumn{3}{c}{Agentic} \\
\cmidrule(lr){2-4}\cmidrule(lr){5-7}
Model & Ens. & Best & $\rho$ & Ens. & Best & $\rho$ \\
\midrule
MLP$^\dagger$ & 5.1618 & 5.2748 & 0.999 & 5.1070 & 5.1981 & 0.997 \\
TabM$^\dagger$ & 5.1751 & 5.2088 & 0.999 & 5.1541 & 5.1925 & 0.998 \\
LightGBM & 5.4697 & 5.4922 & 0.997 & 5.2886 & 5.4848 & 0.996 \\
RealMLP & 5.1536 & 5.2238 & 0.999 & 5.1586 & 5.2889 & 0.999 \\
\midrule
AutoPyTorch & 5.1642 & -- & -- & -- & -- & -- \\
AutoGluon & 5.0869 & -- & -- & 5.0497 & -- & -- \\
\bottomrule
\end{tabular}
\end{minipage}

\end{center}

\Needspace{14\baselineskip}
\begin{center}
\scriptsize
\setlength{\tabcolsep}{2pt}
\renewcommand{\arraystretch}{1.04}

\begin{minipage}[t]{0.49\linewidth}
\centering
\textbf{house ($\times 10^{4}$) $\downarrow$}\\
\begin{tabular}{l@{\hskip 3pt}ccc@{\hskip 5pt}ccc}
\toprule
 & \multicolumn{3}{c}{Default} & \multicolumn{3}{c}{Agentic} \\
\cmidrule(lr){2-4}\cmidrule(lr){5-7}
Model & Ens. & Best & $\rho$ & Ens. & Best & $\rho$ \\
\midrule
MLP$^\dagger$ & 3.0591 & 3.2206 & 0.959 & 3.0086 & 3.1091 & 0.920 \\
TabM$^\dagger$ & 3.0332 & 3.0416 & 0.978 & 2.9927 & 3.0683 & 0.795 \\
LightGBM & 3.1224 & 3.1500 & 0.948 & 3.1270 & 3.1635 & 0.938 \\
RealMLP & 3.0202 & 3.0656 & 0.968 & 2.9877 & 3.0371 & 0.965 \\
\midrule
AutoPyTorch & 3.0247 & -- & -- & -- & -- & -- \\
AutoGluon & 2.9352 & -- & -- & 2.9503 & -- & -- \\
\bottomrule
\end{tabular}
\end{minipage}
\hfill
\begin{minipage}[t]{0.49\linewidth}
\centering
\textbf{microsoft ($\times 10^{-1}$) $\downarrow$}\\
\begin{tabular}{l@{\hskip 3pt}ccc@{\hskip 5pt}ccc}
\toprule
 & \multicolumn{3}{c}{Default} & \multicolumn{3}{c}{Agentic} \\
\cmidrule(lr){2-4}\cmidrule(lr){5-7}
Model & Ens. & Best & $\rho$ & Ens. & Best & $\rho$ \\
\midrule
MLP$^\dagger$ & 7.4381 & 7.4595 & 0.943 & 7.4458 & 7.4930 & 0.929 \\
TabM$^\dagger$ & 7.4217 & 7.4316 & 0.983 & 7.4258 & 7.4535 & 0.941 \\
LightGBM & 7.4217 & 7.4272 & 0.957 & 7.4346 & 7.4371 & 0.981 \\
RealMLP & 7.4413 & 7.4725 & 0.947 & 7.4248 & 7.4382 & 0.943 \\
\midrule
AutoPyTorch & 7.4313 & -- & -- & -- & -- & -- \\
AutoGluon & 7.3968 & -- & -- & 7.3998 & -- & -- \\
\bottomrule
\end{tabular}
\end{minipage}

\end{center}

\Needspace{14\baselineskip}
\begin{center}
\scriptsize
\setlength{\tabcolsep}{2pt}
\renewcommand{\arraystretch}{1.04}

\begin{minipage}[t]{0.49\linewidth}
\centering
\textbf{tabarena/airfoil\_self\_noise $\downarrow$}\\
\begin{tabular}{l@{\hskip 3pt}ccc@{\hskip 5pt}ccc}
\toprule
 & \multicolumn{3}{c}{Default} & \multicolumn{3}{c}{Agentic} \\
\cmidrule(lr){2-4}\cmidrule(lr){5-7}
Model & Ens. & Best & $\rho$ & Ens. & Best & $\rho$ \\
\midrule
MLP$^\dagger$ & 1.0921 & 1.0719 & 0.992 & 1.0495 & 1.2694 & 0.987 \\
TabM$^\dagger$ & 1.0417 & 1.0543 & 0.997 & 0.9682 & 1.0703 & 0.996 \\
LightGBM & 1.6305 & 1.5509 & 0.977 & 1.6208 & 1.8714 & 0.924 \\
RealMLP & 1.0711 & 1.0425 & 0.995 & 1.0546 & 1.1026 & 0.993 \\
\midrule
AutoPyTorch & 1.4413 & -- & -- & -- & -- & -- \\
AutoGluon & 1.1591 & -- & -- & 1.0240 & -- & -- \\
\bottomrule
\end{tabular}
\end{minipage}
\hfill
\begin{minipage}[t]{0.49\linewidth}
\centering
\textbf{tabarena/Amazon\_employee\_access ($\times 10^{-1}$) $\uparrow$}\\
\begin{tabular}{l@{\hskip 3pt}ccc@{\hskip 5pt}ccc}
\toprule
 & \multicolumn{3}{c}{Default} & \multicolumn{3}{c}{Agentic} \\
\cmidrule(lr){2-4}\cmidrule(lr){5-7}
Model & Ens. & Best & $\rho$ & Ens. & Best & $\rho$ \\
\midrule
MLP$^\dagger$ & 8.4789 & 8.3096 & 0.973 & 8.5700 & 8.2742 & 0.882 \\
TabM$^\dagger$ & 8.3749 & 8.2706 & 0.919 & 8.4651 & 8.5256 & -0.106 \\
LightGBM & 8.3321 & 8.1961 & 0.974 & 8.5580 & 8.4597 & 0.957 \\
RealMLP & 8.6927 & 8.5656 & 0.983 & 8.5490 & 8.4874 & 0.953 \\
\midrule
AutoPyTorch & 8.2391 & -- & -- & -- & -- & -- \\
AutoGluon & 8.7546 & -- & -- & 8.7749 & -- & -- \\
\bottomrule
\end{tabular}
\end{minipage}

\end{center}

\Needspace{14\baselineskip}
\begin{center}
\scriptsize
\setlength{\tabcolsep}{2pt}
\renewcommand{\arraystretch}{1.04}

\begin{minipage}[t]{0.49\linewidth}
\centering
\textbf{tabarena/Another-Dataset-on-used-Fiat-500 ($\times 10^{2}$) $\downarrow$}\\
\begin{tabular}{l@{\hskip 3pt}ccc@{\hskip 5pt}ccc}
\toprule
 & \multicolumn{3}{c}{Default} & \multicolumn{3}{c}{Agentic} \\
\cmidrule(lr){2-4}\cmidrule(lr){5-7}
Model & Ens. & Best & $\rho$ & Ens. & Best & $\rho$ \\
\midrule
MLP$^\dagger$ & 7.2749 & 7.4761 & 0.992 & 7.1394 & 7.1956 & 0.978 \\
TabM$^\dagger$ & 7.2183 & 7.2410 & 0.997 & 7.2735 & 7.2945 & 0.970 \\
LightGBM & 7.2233 & 7.1797 & 0.965 & 7.1508 & 7.0806 & 0.970 \\
RealMLP & 6.9871 & 7.1376 & 0.988 & 7.2087 & 7.5569 & 0.984 \\
\midrule
AutoPyTorch & 7.0074 & -- & -- & -- & -- & -- \\
AutoGluon & 6.9228 & -- & -- & 6.9843 & -- & -- \\
\bottomrule
\end{tabular}
\end{minipage}
\hfill
\begin{minipage}[t]{0.49\linewidth}
\centering
\textbf{tabarena/APSFailure ($\times 10^{-1}$) $\uparrow$}\\
\begin{tabular}{l@{\hskip 3pt}ccc@{\hskip 5pt}ccc}
\toprule
 & \multicolumn{3}{c}{Default} & \multicolumn{3}{c}{Agentic} \\
\cmidrule(lr){2-4}\cmidrule(lr){5-7}
Model & Ens. & Best & $\rho$ & Ens. & Best & $\rho$ \\
\midrule
MLP$^\dagger$ & 9.9388 & 9.9243 & 0.996 & 9.9531 & 9.8932 & 0.986 \\
TabM$^\dagger$ & 9.9400 & 9.9317 & 0.996 & 9.9543 & 9.9250 & 0.992 \\
LightGBM & 9.8816 & 9.9071 & 0.983 & 9.9428 & 9.9421 & 0.982 \\
RealMLP & 9.9254 & 9.9397 & 0.997 & 9.9404 & 9.9328 & 0.996 \\
\midrule
AutoPyTorch & 9.9338 & -- & -- & -- & -- & -- \\
AutoGluon & 9.9617 & -- & -- & 9.9509 & -- & -- \\
\bottomrule
\end{tabular}
\end{minipage}

\end{center}

\Needspace{14\baselineskip}
\begin{center}
\scriptsize
\setlength{\tabcolsep}{2pt}
\renewcommand{\arraystretch}{1.04}

\begin{minipage}[t]{0.49\linewidth}
\centering
\textbf{tabarena/bank-marketing ($\times 10^{-1}$) $\uparrow$}\\
\begin{tabular}{l@{\hskip 3pt}ccc@{\hskip 5pt}ccc}
\toprule
 & \multicolumn{3}{c}{Default} & \multicolumn{3}{c}{Agentic} \\
\cmidrule(lr){2-4}\cmidrule(lr){5-7}
Model & Ens. & Best & $\rho$ & Ens. & Best & $\rho$ \\
\midrule
MLP$^\dagger$ & 7.7486 & 7.7187 & 0.996 & 7.7263 & 7.6759 & 0.939 \\
TabM$^\dagger$ & 7.7408 & 7.7406 & 0.998 & 7.7250 & 7.7256 & 0.994 \\
LightGBM & 7.6496 & 7.5439 & 0.958 & 7.6322 & 7.5159 & 0.976 \\
RealMLP & 7.7214 & 7.6891 & 0.995 & 7.7328 & 7.7042 & 0.994 \\
\midrule
AutoPyTorch & 7.7188 & -- & -- & -- & -- & -- \\
AutoGluon & 7.7255 & -- & -- & 7.7536 & -- & -- \\
\bottomrule
\end{tabular}
\end{minipage}
\hfill
\begin{minipage}[t]{0.49\linewidth}
\centering
\textbf{tabarena/Bioresponse ($\times 10^{-1}$) $\uparrow$}\\
\begin{tabular}{l@{\hskip 3pt}ccc@{\hskip 5pt}ccc}
\toprule
 & \multicolumn{3}{c}{Default} & \multicolumn{3}{c}{Agentic} \\
\cmidrule(lr){2-4}\cmidrule(lr){5-7}
Model & Ens. & Best & $\rho$ & Ens. & Best & $\rho$ \\
\midrule
MLP$^\dagger$ & 8.6307 & 8.5387 & 0.866 & 8.7010 & 8.6130 & 0.774 \\
TabM$^\dagger$ & 8.7630 & 8.7514 & 0.968 & 8.7947 & 8.6292 & 0.349 \\
LightGBM & 8.8308 & 8.7361 & 0.668 & 8.8487 & 8.8535 & 0.495 \\
RealMLP & 8.7836 & 8.5722 & 0.878 & 8.7568 & 8.6600 & 0.793 \\
\midrule
AutoPyTorch & 8.7056 & -- & -- & -- & -- & -- \\
AutoGluon & 8.8883 & -- & -- & 8.8907 & -- & -- \\
\bottomrule
\end{tabular}
\end{minipage}

\end{center}

\Needspace{14\baselineskip}
\begin{center}
\scriptsize
\setlength{\tabcolsep}{2pt}
\renewcommand{\arraystretch}{1.04}

\begin{minipage}[t]{0.49\linewidth}
\centering
\textbf{tabarena/concrete\_compressive\_strength $\downarrow$}\\
\begin{tabular}{l@{\hskip 3pt}ccc@{\hskip 5pt}ccc}
\toprule
 & \multicolumn{3}{c}{Default} & \multicolumn{3}{c}{Agentic} \\
\cmidrule(lr){2-4}\cmidrule(lr){5-7}
Model & Ens. & Best & $\rho$ & Ens. & Best & $\rho$ \\
\midrule
MLP$^\dagger$ & 4.5909 & 4.6376 & 0.989 & 4.1806 & 4.1483 & 0.826 \\
TabM$^\dagger$ & 4.4148 & 4.6771 & 0.984 & 4.4071 & 4.7297 & 0.978 \\
LightGBM & 4.6905 & 4.6727 & 0.978 & 4.4771 & 4.5654 & 0.972 \\
RealMLP & 4.6129 & 4.9984 & 0.987 & 4.4865 & 4.9832 & 0.707 \\
\midrule
AutoPyTorch & 4.5269 & -- & -- & -- & -- & -- \\
AutoGluon & 4.1461 & -- & -- & 3.9968 & -- & -- \\
\bottomrule
\end{tabular}
\end{minipage}
\hfill
\begin{minipage}[t]{0.49\linewidth}
\centering
\textbf{tabarena/credit\_card\_clients\_default ($\times 10^{-1}$) $\uparrow$}\\
\begin{tabular}{l@{\hskip 3pt}ccc@{\hskip 5pt}ccc}
\toprule
 & \multicolumn{3}{c}{Default} & \multicolumn{3}{c}{Agentic} \\
\cmidrule(lr){2-4}\cmidrule(lr){5-7}
Model & Ens. & Best & $\rho$ & Ens. & Best & $\rho$ \\
\midrule
MLP$^\dagger$ & 7.8881 & 7.8898 & 0.991 & 7.8919 & 7.8699 & 0.844 \\
TabM$^\dagger$ & 7.9085 & 7.9051 & 0.998 & 7.9266 & 7.8867 & 0.982 \\
LightGBM & 7.8757 & 7.8719 & 0.931 & 7.8643 & 7.8703 & 0.792 \\
RealMLP & 7.8777 & 7.8919 & 0.977 & 7.9084 & 7.9152 & 0.979 \\
\midrule
AutoPyTorch & 7.8629 & -- & -- & -- & -- & -- \\
AutoGluon & 7.9133 & -- & -- & 7.9356 & -- & -- \\
\bottomrule
\end{tabular}
\end{minipage}

\end{center}

\Needspace{14\baselineskip}
\begin{center}
\scriptsize
\setlength{\tabcolsep}{2pt}
\renewcommand{\arraystretch}{1.04}

\begin{minipage}[t]{0.49\linewidth}
\centering
\textbf{tabarena/customer\_satisfaction\_in\_airline ($\times 10^{-1}$) $\uparrow$}\\
\begin{tabular}{l@{\hskip 3pt}ccc@{\hskip 5pt}ccc}
\toprule
 & \multicolumn{3}{c}{Default} & \multicolumn{3}{c}{Agentic} \\
\cmidrule(lr){2-4}\cmidrule(lr){5-7}
Model & Ens. & Best & $\rho$ & Ens. & Best & $\rho$ \\
\midrule
MLP$^\dagger$ & 9.9456 & 9.9414 & 0.986 & 9.9462 & 9.9381 & 0.974 \\
TabM$^\dagger$ & 9.9519 & 9.9512 & 0.997 & 9.9527 & 9.9514 & 0.459 \\
LightGBM & 9.9418 & 9.9420 & 0.832 & 9.9425 & 9.9425 & 0.991 \\
RealMLP & 9.9507 & 9.9479 & 0.993 & 9.9517 & 9.9496 & 0.991 \\
\midrule
AutoPyTorch & 9.9444 & -- & -- & -- & -- & -- \\
AutoGluon & 9.9553 & -- & -- & 9.9547 & -- & -- \\
\bottomrule
\end{tabular}
\end{minipage}
\hfill
\begin{minipage}[t]{0.49\linewidth}
\centering
\textbf{tabarena/diabetes ($\times 10^{-1}$) $\uparrow$}\\
\begin{tabular}{l@{\hskip 3pt}ccc@{\hskip 5pt}ccc}
\toprule
 & \multicolumn{3}{c}{Default} & \multicolumn{3}{c}{Agentic} \\
\cmidrule(lr){2-4}\cmidrule(lr){5-7}
Model & Ens. & Best & $\rho$ & Ens. & Best & $\rho$ \\
\midrule
MLP$^\dagger$ & 8.2550 & 8.1754 & 0.783 & 8.3527 & 8.4016 & 0.585 \\
TabM$^\dagger$ & 7.5817 & 7.5783 & 0.986 & 8.2483 & 8.2584 & 0.743 \\
LightGBM & 8.1991 & 7.9973 & 0.999 & 7.9227 & 7.2828 & 1.000 \\
RealMLP & 8.4357 & 8.4083 & 0.920 & 8.2811 & 8.3447 & 0.919 \\
\midrule
AutoPyTorch & 8.1566 & -- & -- & -- & -- & -- \\
AutoGluon & 8.4076 & -- & -- & 8.3835 & -- & -- \\
\bottomrule
\end{tabular}
\end{minipage}

\end{center}

\Needspace{14\baselineskip}
\begin{center}
\scriptsize
\setlength{\tabcolsep}{2pt}
\renewcommand{\arraystretch}{1.04}

\begin{minipage}[t]{0.49\linewidth}
\centering
\textbf{tabarena/Diabetes130US ($\times 10^{-1}$) $\uparrow$}\\
\begin{tabular}{l@{\hskip 3pt}ccc@{\hskip 5pt}ccc}
\toprule
 & \multicolumn{3}{c}{Default} & \multicolumn{3}{c}{Agentic} \\
\cmidrule(lr){2-4}\cmidrule(lr){5-7}
Model & Ens. & Best & $\rho$ & Ens. & Best & $\rho$ \\
\midrule
MLP$^\dagger$ & 6.6519 & 6.6647 & 0.998 & 6.7481 & 6.6587 & 0.986 \\
TabM$^\dagger$ & 6.7300 & 6.7372 & 1.000 & 6.7645 & 6.6671 & 0.997 \\
LightGBM & 6.6741 & 6.6575 & 0.989 & 6.7327 & 6.6237 & 0.921 \\
RealMLP & 6.7327 & 6.7111 & 0.990 & 6.7642 & 6.7286 & 0.979 \\
\midrule
AutoPyTorch & 6.7263 & -- & -- & -- & -- & -- \\
AutoGluon & 6.8475 & -- & -- & 6.8117 & -- & -- \\
\bottomrule
\end{tabular}
\end{minipage}
\hfill
\begin{minipage}[t]{0.49\linewidth}
\centering
\textbf{tabarena/E-CommereShippingData ($\times 10^{-1}$) $\uparrow$}\\
\begin{tabular}{l@{\hskip 3pt}ccc@{\hskip 5pt}ccc}
\toprule
 & \multicolumn{3}{c}{Default} & \multicolumn{3}{c}{Agentic} \\
\cmidrule(lr){2-4}\cmidrule(lr){5-7}
Model & Ens. & Best & $\rho$ & Ens. & Best & $\rho$ \\
\midrule
MLP$^\dagger$ & 7.3964 & 7.4121 & 0.939 & 7.3701 & 7.3701 & -- \\
TabM$^\dagger$ & 7.3678 & 7.3547 & 0.995 & 7.3983 & 7.4361 & 0.502 \\
LightGBM & 7.3731 & 7.3401 & 0.694 & 7.4145 & 7.4145 & -0.012 \\
RealMLP & 7.4131 & 7.3659 & 0.971 & 7.4248 & 7.4181 & 0.939 \\
\midrule
AutoPyTorch & 7.4104 & -- & -- & -- & -- & -- \\
AutoGluon & 7.4040 & -- & -- & 7.4084 & -- & -- \\
\bottomrule
\end{tabular}
\end{minipage}

\end{center}

\Needspace{14\baselineskip}
\begin{center}
\scriptsize
\setlength{\tabcolsep}{2pt}
\renewcommand{\arraystretch}{1.04}

\begin{minipage}[t]{0.49\linewidth}
\centering
\textbf{tabarena/Food\_Delivery\_Time $\downarrow$}\\
\begin{tabular}{l@{\hskip 3pt}ccc@{\hskip 5pt}ccc}
\toprule
 & \multicolumn{3}{c}{Default} & \multicolumn{3}{c}{Agentic} \\
\cmidrule(lr){2-4}\cmidrule(lr){5-7}
Model & Ens. & Best & $\rho$ & Ens. & Best & $\rho$ \\
\midrule
MLP$^\dagger$ & 7.6165 & 7.6147 & 0.936 & 7.4622 & 7.4826 & 0.935 \\
TabM$^\dagger$ & 7.6896 & 7.6884 & 0.944 & 7.4060 & 7.4684 & 0.937 \\
LightGBM & 7.5217 & 7.5367 & 0.899 & 7.3161 & 7.3560 & 0.856 \\
RealMLP & 7.3778 & 7.3825 & 0.951 & 7.5389 & 7.6036 & 0.924 \\
\midrule
AutoPyTorch & 7.4385 & -- & -- & -- & -- & -- \\
AutoGluon & 7.4264 & -- & -- & 7.5950 & -- & -- \\
\bottomrule
\end{tabular}
\end{minipage}
\hfill
\begin{minipage}[t]{0.49\linewidth}
\centering
\textbf{tabarena/GiveMeSomeCredit ($\times 10^{-1}$) $\uparrow$}\\
\begin{tabular}{l@{\hskip 3pt}ccc@{\hskip 5pt}ccc}
\toprule
 & \multicolumn{3}{c}{Default} & \multicolumn{3}{c}{Agentic} \\
\cmidrule(lr){2-4}\cmidrule(lr){5-7}
Model & Ens. & Best & $\rho$ & Ens. & Best & $\rho$ \\
\midrule
MLP$^\dagger$ & 8.7073 & 8.6997 & 0.999 & 8.7016 & 8.6879 & 0.988 \\
TabM$^\dagger$ & 8.7062 & 8.6994 & 0.999 & 8.7122 & 8.7081 & 0.994 \\
LightGBM & 8.6910 & 8.6881 & 0.955 & 8.6628 & 8.6756 & 0.740 \\
RealMLP & 8.6931 & 8.6834 & 0.999 & 8.6925 & 8.6911 & 0.985 \\
\midrule
AutoPyTorch & 8.6934 & -- & -- & -- & -- & -- \\
AutoGluon & 8.7078 & -- & -- & 8.7084 & -- & -- \\
\bottomrule
\end{tabular}
\end{minipage}

\end{center}

\Needspace{14\baselineskip}
\begin{center}
\scriptsize
\setlength{\tabcolsep}{2pt}
\renewcommand{\arraystretch}{1.04}

\begin{minipage}[t]{0.49\linewidth}
\centering
\textbf{tabarena/healthcare\_insurance\_expenses ($\times 10^{3}$) $\downarrow$}\\
\begin{tabular}{l@{\hskip 3pt}ccc@{\hskip 5pt}ccc}
\toprule
 & \multicolumn{3}{c}{Default} & \multicolumn{3}{c}{Agentic} \\
\cmidrule(lr){2-4}\cmidrule(lr){5-7}
Model & Ens. & Best & $\rho$ & Ens. & Best & $\rho$ \\
\midrule
MLP$^\dagger$ & 4.2428 & 4.4211 & 0.992 & 4.2734 & 4.6018 & 0.988 \\
TabM$^\dagger$ & 4.1157 & 4.1540 & 0.995 & 4.2239 & 4.2813 & 0.992 \\
LightGBM & 4.2900 & 4.2384 & 0.990 & 4.3421 & 4.3323 & 0.964 \\
RealMLP & 4.1561 & 4.1509 & 0.994 & 4.1224 & 4.1451 & 0.993 \\
\midrule
AutoPyTorch & 4.3118 & -- & -- & -- & -- & -- \\
AutoGluon & 4.0551 & -- & -- & 4.0598 & -- & -- \\
\bottomrule
\end{tabular}
\end{minipage}
\hfill
\begin{minipage}[t]{0.49\linewidth}
\centering
\textbf{tabarena/heloc ($\times 10^{-1}$) $\uparrow$}\\
\begin{tabular}{l@{\hskip 3pt}ccc@{\hskip 5pt}ccc}
\toprule
 & \multicolumn{3}{c}{Default} & \multicolumn{3}{c}{Agentic} \\
\cmidrule(lr){2-4}\cmidrule(lr){5-7}
Model & Ens. & Best & $\rho$ & Ens. & Best & $\rho$ \\
\midrule
MLP$^\dagger$ & 7.9570 & 7.9454 & 0.974 & 7.9828 & 7.9527 & 0.394 \\
TabM$^\dagger$ & 7.9660 & 7.9696 & 0.992 & 7.9823 & 7.9663 & 0.526 \\
LightGBM & 7.9473 & 7.9340 & 0.743 & 7.9429 & 7.9444 & 0.036 \\
RealMLP & 7.9982 & 7.9770 & 0.954 & 7.9889 & 7.9830 & 0.950 \\
\midrule
AutoPyTorch & 7.9512 & -- & -- & -- & -- & -- \\
AutoGluon & 7.9772 & -- & -- & 7.9812 & -- & -- \\
\bottomrule
\end{tabular}
\end{minipage}

\end{center}

\Needspace{14\baselineskip}
\begin{center}
\scriptsize
\setlength{\tabcolsep}{2pt}
\renewcommand{\arraystretch}{1.04}

\begin{minipage}[t]{0.49\linewidth}
\centering
\textbf{tabarena/HR\_Analytics\_Job\_Change\_of\_Data\_Scientists ($\times 10^{-1}$) $\uparrow$}\\
\begin{tabular}{l@{\hskip 3pt}ccc@{\hskip 5pt}ccc}
\toprule
 & \multicolumn{3}{c}{Default} & \multicolumn{3}{c}{Agentic} \\
\cmidrule(lr){2-4}\cmidrule(lr){5-7}
Model & Ens. & Best & $\rho$ & Ens. & Best & $\rho$ \\
\midrule
MLP$^\dagger$ & 8.1123 & 8.0956 & 0.986 & 8.1218 & 8.0863 & 0.620 \\
TabM$^\dagger$ & 8.1293 & 8.1237 & 0.993 & 8.1067 & 8.0417 & 0.981 \\
LightGBM & 8.0766 & 8.0706 & 0.854 & 8.0970 & 8.1152 & 0.786 \\
RealMLP & 8.1128 & 8.1023 & 0.987 & 8.1219 & 8.0735 & 0.984 \\
\midrule
AutoPyTorch & 8.0876 & -- & -- & -- & -- & -- \\
AutoGluon & 8.1381 & -- & -- & 8.1415 & -- & -- \\
\bottomrule
\end{tabular}
\end{minipage}
\hfill
\begin{minipage}[t]{0.49\linewidth}
\centering
\textbf{tabarena/in\_vehicle\_coupon\_recommendation ($\times 10^{-1}$) $\uparrow$}\\
\begin{tabular}{l@{\hskip 3pt}ccc@{\hskip 5pt}ccc}
\toprule
 & \multicolumn{3}{c}{Default} & \multicolumn{3}{c}{Agentic} \\
\cmidrule(lr){2-4}\cmidrule(lr){5-7}
Model & Ens. & Best & $\rho$ & Ens. & Best & $\rho$ \\
\midrule
MLP$^\dagger$ & 8.3507 & 8.2174 & 0.794 & 8.4306 & 8.3132 & 0.743 \\
TabM$^\dagger$ & 8.5539 & 8.5323 & 0.905 & 8.5120 & 8.5156 & 0.530 \\
LightGBM & 8.4024 & 8.4187 & 0.610 & 8.3935 & 8.3868 & -0.319 \\
RealMLP & 8.4935 & 8.4211 & 0.927 & 8.5175 & 8.4445 & 0.885 \\
\midrule
AutoPyTorch & 8.4038 & -- & -- & -- & -- & -- \\
AutoGluon & 8.5373 & -- & -- & 8.5634 & -- & -- \\
\bottomrule
\end{tabular}%
\end{minipage}

\end{center}

\section{TabArena Per-Dataset Results}
\label{app:tabarena-per-dataset}

Table~\ref{tab:tabarena_per_dataset_results} reports the per-dataset results behind the TabArena-Lite comparison in Appendix~\ref{app:tabarena}. Each subtable lists the five model families and reports, for the base and the agentic (-A) search space, the score of the default configuration (Def.), of the best of the $200$ random configurations by validation score (Tuned), and of the greedy ensemble built from the same configurations (T+E). Scores use the official TabArena metric of each dataset: AUROC for binary classification, log-loss for multiclass, and RMSE for regression. Base rows are the official TabArena results, except MLP$^\dagger$, which has no official TabArena counterpart and whose base row is our own run under the same protocol, sharing its default configuration with MLP$^\dagger$-A. The official TabICLv2 entry is default-only, so its Tuned and T+E cells are marked as missing.

\Needspace{18\baselineskip}
\begin{center}
\captionof{table}{Per-dataset TabArena-Lite results for the 51 datasets: default (Def.), tuned (Tuned), and tuned-plus-ensembled (T+E) test scores for the base and agentic search space of each model family. Arrows indicate whether higher or lower values are better.}
\label{tab:tabarena_per_dataset_results}
\end{center}

\Needspace{16\baselineskip}
\begin{center}
\scriptsize
\setlength{\tabcolsep}{2pt}
\renewcommand{\arraystretch}{1.04}

\begin{minipage}[t]{0.49\linewidth}
\centering
\textbf{airfoil\_self\_noise, RMSE $\downarrow$}\\

\end{minipage}

\end{center}

\end{document}